\documentclass[]{interact}

\usepackage{epstopdf}
\usepackage[caption=false]{subfig}

\usepackage{subcaption}

\usepackage{graphicx}%
\usepackage{multirow}%
\usepackage{amsmath,amssymb,amsfonts}%
\usepackage{amsthm}%
\usepackage{mathrsfs}%
\usepackage[title]{appendix}%
\usepackage{xcolor}%
\usepackage{textcomp}%
\usepackage{manyfoot}%
\usepackage{booktabs}%
\usepackage{algorithm}%
\usepackage{algorithmicx}%
\usepackage{algpseudocode}%
\usepackage{listings}
\usepackage[numbers,sort&compress]{natbib}
\bibpunct[, ]{[}{]}{,}{n}{,}{,}
\renewcommand\bibfont{\fontsize{10}{12}\selectfont}
\makeatletter
\def\NAT@def@citea{\def\@citea{\NAT@separator}}
\makeatother

\theoremstyle{plain}

\theoremstyle{definition}

\theoremstyle{remark}

\begin{document}

\title{A Novel Narrow Region Detector for Sampling-Based Planners' Efficiency: Match Based Passage Identifier}

\author{
\name{Yafes Enes Şahiner\textsuperscript{a}, Esat Yusuf Gündoğdu\textsuperscript{a} and Volkan Sezer\textsuperscript{a}\thanks{CONTACT Yafes Enes Şahiner Email: sahiner18@itu.edu.tr}}
\affil{\textsuperscript{a}Smart and Autonomous Systems Lab., Istanbul Technical University, Istanbul, Türkiye}
}

\maketitle

\begin{abstract}
Autonomous technology, which has become widespread today, appears in many different configurations such as mobile robots, manipulators, and drones. One of the most important tasks of these vehicles during autonomous operations is path planning. In the literature, path planners are generally divided into two categories: probabilistic and deterministic methods. In the analysis of probabilistic methods, the common problem of almost all methods is observed in narrow passage environments. In this paper, a novel sampler is proposed that deterministically identifies narrow passage environments using occupancy grid maps and accordingly increases the amount of sampling in these regions. The codes of the algorithm is provided as open source. To evaluate the performance of the algorithm, benchmark studies are conducted in three distinct categories: specific and random simulation environments, and a real-world environment. As a result, it is observed that our algorithm provides higher performance in planning time and number of milestones compared to the baseline samplers. 
\end{abstract}


\section{Introduction}\label{sec:introduction}
Path planning is a core component of autonomous systems. Path planning methods for robotic systems have been traditionally approached through two fundamental methodologies: deterministic and probabilistic.  A* \citep{hart1968formal}, Dijkstra \citep{dijkstra2022note}, JPS \citep{harabor2011online}, and Visibility Graph are some of the deterministic planners. These approaches are classified as deterministic due to their consistency in providing identical solutions to identical problems. In addition to deterministic methods, sampling-based methods like PRM (Probabilistic Roadmap) \citep{kavraki1996probabilistic} and RRT (Rapidly-exploring Random Trees) \citep{lavalle1998rapidly}, have been developed. These methods provide different solutions to the identical problem each time. However, these sampling-based algorithms have demonstrated efficiency, particularly in high-dimensional configuration spaces like robotic manipulators. A detailed description and comparative review of sampling-based algorithms can be found in \citep{orthey2023sampling}.

PRM consists of two parts: the build phase and the query phase. In the build phase, the configuration space is sampled independently from the initial and target points, then these nodes are connected with a suitable local planner to create a roadmap. The k-nearest neighbor approach can be used to determine the nodes to be connected. Each node is linked with a certain number of nodes nearest to it.  This stage creates a roadmap that can be used for different initial and target points. However, the roadmap must be reconstructed or modified at certain time intervals in dynamic enviroments. In the query phase, the path is calculated on the roadmap with a graph solver such as A* or Dijkstra.

\begin{figure}
\centering
\includegraphics[width=9cm]{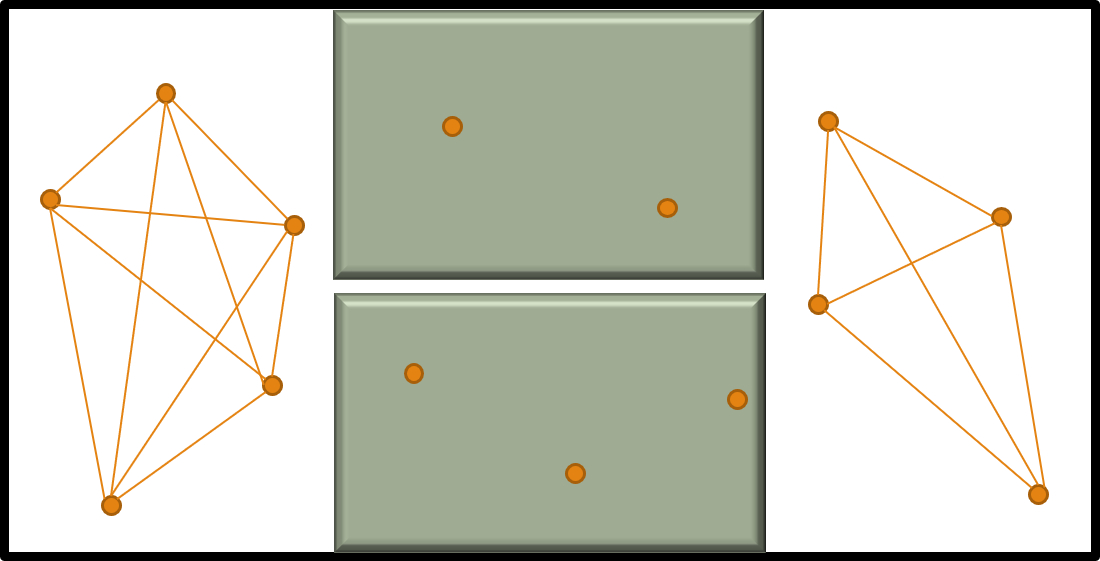}
\caption{Connectivity problem in narrow passage environment}
\label{fig:connectivity-problem}
\end{figure}

There are two characteristics that evaluate the effectiveness of PRM roadmaps: coverage and connectivity \citep{rantanen2014construct}. Coverage refers to whether the roadmap is spread over the entire map. If there is a roadmap with high coverage, a new sample can be directly connected to roadmap with a local planner. Connectivity is the ability to calculate a path between any two samples selected from roadmap. This means that the number of subgraphs is low in the roadmap with strong connectivity. Although PRM is generally efficient, its connectivity may not be ideal in environments with narrow passages as shown in Figure \ref{fig:connectivity-problem}. Because when uniform sampling is performed the probability of sampling narrow passages is poor due to the small volume they occupy in the map. In such environments, both planning time and number of milestones of PRM are much higher.

To solve these drawbacks, narrow regions need to be prioritized. If narrow regions can be identified, PRM performance can be improved by sampling the relevant regions. In \citep{yershova2005dynamic}, it is stated that narrow passage detection is an more challenging task than motion planning. In this paper, a novel method that deterministically detects narrow regions using an occupancy grid map is presented. This method aims to improve connectivity of probabilistic roadmaps by weighted sampling of detected narrow passages according to their narrowness. However, it is not possible to produce roadmaps with good coverage by only sampling narrow passages. Therefore, in addition to narrow passage sampler, uniform sampler is used in a hybrid approach. As a result, roadmaps that have strong coverage and connectivity properties are obtained. We have tested our work in both simulation and real-world environments and compared it with other methods. As a result of the experiments, it has been observed that our algorithm is outperforming other methods in terms of planning time and number of milestones in roadmap because milestones generated by other samplers are not guaranteed to increase connectivity of roadmaps. Sampling around obstacles or dead-end regions causes more samples to calculate a valid path, which leads to worse results in terms of planning time and number of milestones. In addition, path length is usually increased as a result of the generation of a path over the samples in these areas. On the other hand, Match Based Passage Identifier directly detects and samples narrow regions, so that it provides samples that increase connectivity of roadmaps.

To summarise, the major contributions of this paper are as follows:
\begin{itemize}
    \item A novel algorithm that deterministically detects narrow passages is presented.
    \item A weighted sampler is proposed that generates milestones according to the narrowness of the regions on the map.
    \item The method is tested on custom and randomly generated maps and compared in the open source Open Motion Planning Library (OMPL) environment \citep{sucan2012open}.
    \item The real-time applicability of the algorithm is shown through real-world experiments.
    \item The implementation codes of the algorithm is shared as open source \citep{git-repo}.
\end{itemize}

The rest of the paper is organized as follows. Section \ref{sec:literature-review} presents the related works on narrow passage problem. Section \ref{sec:technical-approach} describes the technical details of the method. Section \ref{sec:experiments} compares the performances of our algorithm and other methods with experiments. Section \ref{sec:conclusion-and-future-work} concludes the paper and summarises the results. In addition, future work is provided for the further development of the study.

\section{Literature Review}\label{sec:literature-review}
Various methods have been developed to prevent narrow passage problem encountered when using probabilistic roadmap method. The common point of almost all methods is to provide a connection on the narrow passage with a sampler that increases number of milestones in these areas.

Obstacle Based PRM (OBPRM) \citep{amato1998obprm} is a sampler that aims to increase the number of samples at the boundaries of obstacles. For each sampling, a random point and a random direction are selected in the environment. Collision checking is performed from a random point until an obstacle is encountered in a random direction. The point before the collision occurs is considered as the milestone in the graph. Uniform Obstacle Based PRM (UOBPRM) \citep{yeh2012uobprm} is a variant of OBPRM. In addition to sampling at the boundaries of the obstacles, is to distribute the samples homogeneously within the boundaries of the obstacles. To achieve this, line segments of a certain length are randomly placed on the map and a collision check is performed on this line segment at a certain interval of steps. With each validity change, it is added to the graph as a milestone. Performing additional collision checks during milestone determination for both of these increases the sampling time.

The method that can sample close to the obstacle boundaries without performing additional collision checks is Gaussian Sampler \citep{boor1999gaussian}. This method selects a random point for each sampling operation and a second point within a certain distance from the first point. If one of the two points is valid and the other is invalid, the valid point is added to the graph. Creating milestones at obstacle boundaries is similarly seen in retraction-based methods. Invalid samples are moved to the contact surface. In \citep{saha2005finding}, the aim is to move the invalid sample to the nearest valid point. For this, the obstacle surface is sampled until the point that minimizes the distance locally is found. All sampled points are included in the graph. The methods described in \citep{amato1998obprm}, \citep{yeh2012uobprm}, \citep{boor1999gaussian}, and \citep{saha2005finding} are intended for sampling at obstacle boundaries, as stated. Planning through these regions is not the best for the safety of the moving robot. 

On the other hand, Medial Axis PRM (MAPRM) \citep{wilmarth1999maprm} is a planner that gives importance to safety and creates a route away from obstacles. First, it calculates the medial axis of the environment and then makes samples on the medial axis. However, calculating the medial axis consumes a certain amount of time. Although the methods described in \citep{amato1998obprm}, \citep{yeh2012uobprm}, \citep{boor1999gaussian} and \citep{wilmarth1999maprm} increase the number of samples in narrow passages compared to the uniform sampler, they are not intended for sampling directly on the narrow passages. 

Randomized Bridge Builder (RBB) \citep{sun2005narrow} is developed to provide intensive sampling within narrow passages. The method, known as the Bridge Test, uses three samples to create each milestone. Certain conditions must be met to create a milestone. The first sample should be on the obstacle, the second sample selected close to the first sample should be on the obstacle, and the third sample, calculated as the midpoint of the first two samples, should be on free space. RBB creates a milestone by doing with three samples what the Gaussian sampler does with two samples. Even though it samples more, it manages to concentrate more on narrow passages. The downside of the Bridge test is that it cannot distinguish between dead-end regions and narrow passages because it analyzes a local area of the map. In addition, Bridge Test is used in different planners. One of these is Learning Multi RRT (LM-RRT) \citep{wang2018learning}. This method samples repeatedly using RBB. Then, it clusters these samples and starts an RRT tree in each cluster. The selection of the tree to be grown is analogized to the multi-armed bandit problem. This problem is handled by $\epsilon_t$-greedy, which is a reinforcement learning method.

Furthermore, there are different planners that address narrow passage problem with RRT methods. One of these is Spark PRM \citep{shi2014spark}, which uses PRM and RRT hybridly. In this method, samples are made initially using PRM, and a graph is created. If a connectivity problem occurs, subgraphs that are connected with a low number of milestones are observed. In this case, Spark PRM starts RRT from a point belonging to the subgraph consisting of a small number of milestones. Thanks to the growing RRT tree, it is aimed to improve connectivity with other graphs. Additionally, a utility-guided approach to RRT \citep{burns2007single} is proposed, where the expansion strategy is adapted to maximize expected utility, which depends on the selected nodes, their expansion direction, and expansion distance. Moreover, different methods are combined with PRM. An example of this is the hybrid use of PRM with the approximate cell decomposition \citep{zhang2007hybrid}. It is aimed to ensure the connection of semi-filled cells with each other with PRM milestones. Thus, it is aimed to improve the connection feature in environments with narrow passages. Another planner that aims to increase sampling in narrow passage areas is the Toggle PRM \citep{denny2011toggle}. In this method, two graphs are created, one of these grows on obstacles and the other in free space. While one of the graphs is growing, if an invalid sample is detected, it is added to the other graph. 

Narrow passage problem is seen in multi-level motion planning. The method \citep{orthey2021section} is developed for high-dimensional motion planning problems by first projecting the configuration space into lower dimensions to obtain a base path. This base path then serves as an admissible heuristic for the complex robot model, allowing for efficient exploration of narrow passages. In addition, a method based on Minkowski sum \citep{ruan2022efficient} utilizes unions of ellipsoids to parameterize C-space obstacles in closed form, allowing the use of prior knowledge to avoid sampling invalid configurations for high-dimensional motion planning.

There are learning-based approaches to solve narrow passage problem. Sample Driven Connectivity Learning (SDCL) \citep{sampledriven2023} is a new learning-based method for narrow passage problem. In this method, firstly, the regions that make PRM roadmaps difficult to connect are learned and then these regions are sampled. In the learning phase, the samples are divided into two the ones connected to goal point and the others, and then the binary classifier is trained with support vector machine. In the sampling stage, the region separating two classes is sampled and this process is repeated until a valid path is found, thus accelerating the connection process. Moreover, the method \citep{ichter2018learning} focuses on learning sampling distributions. It uses a conditional variational autoencoder (CVAE) to sample from the latent space, conditioned on the specifics of planning problem, and prioritizes regions where optimal motion plans are likely to be found.

There are many studies on the solution of narrow passage problem. To make a fair comparison between our method and existing methods, OMPL is used. However, OMPL does not contain the implementation of all the algorithms described. The benchmark is performed with the Bridge Test, Gaussian, Obstacle-based, and Uniform samplers available in OMPL. Section \ref{sec:experiments} provides the details and results of the experiments.

\section{Technical Approach}\label{sec:technical-approach}
In this section, the technical details of the Match Based Passage Identifier that detects the narrow passages in grid-based environments are given. Then, the details of the weighted sampling on the detected narrow passages are explained. 

\subsection{Passage Identification Algorithm}\label{sec:passage-identification-algorithm}

\begin{figure}
\centering
\subfloat[Occupancy grid map]{%
\label{fig:compDef1}
\includegraphics[width=0.35\textwidth]{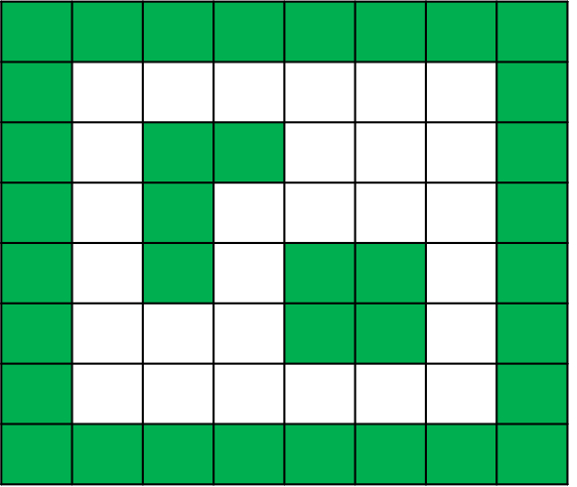}}
\hspace{10pt}
\subfloat[Obstacle components]{%
\label{fig:compDef2}
\includegraphics[width=0.35\textwidth]{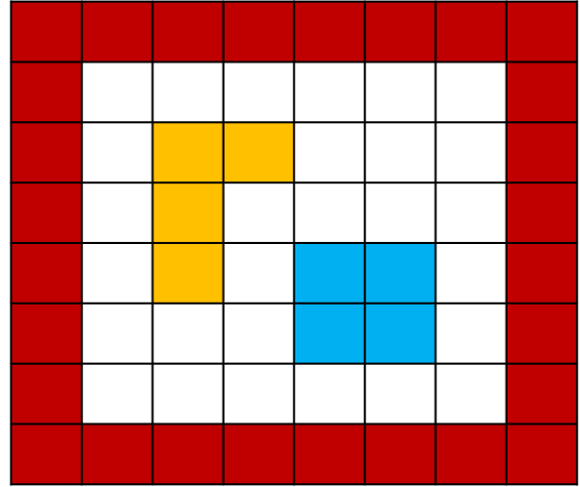}}
\caption{Clustering of obstacle units}
\label{fig:compDef}
\end{figure}

Match Based Passage Identifier comprises four steps: finding connected components, calculating foreign matches, calculating self-matches, and collision checking of matches. These stages are outlined in the pseudocode presented in Algorithm \ref{algo1}. To understand the algorithm, it is essential to clarify the terminology employed. The process operates on an occupancy grid map, where every cell shown as empty is termed a \textit{free space unit}. These free space units, when adjacent to each other, form a structure called a \textit{free space component}. Similarly, each cell marked as occupied on the map is referred to as a \textit{obstacle unit}, and the structure resulting from the adjacency of such obstacle units is referred to as a \textit{obstacle component}. For instance, the occupancy grid map shown in Figure \ref{fig:compDef1} is clustered, with the resulting obstacle components shown in different colors in Figure \ref{fig:compDef2}.

\begin{figure}
\centering
\subfloat[Foreign matches]{%
\label{fig:foreign-match}
\includegraphics[width=0.35\textwidth]{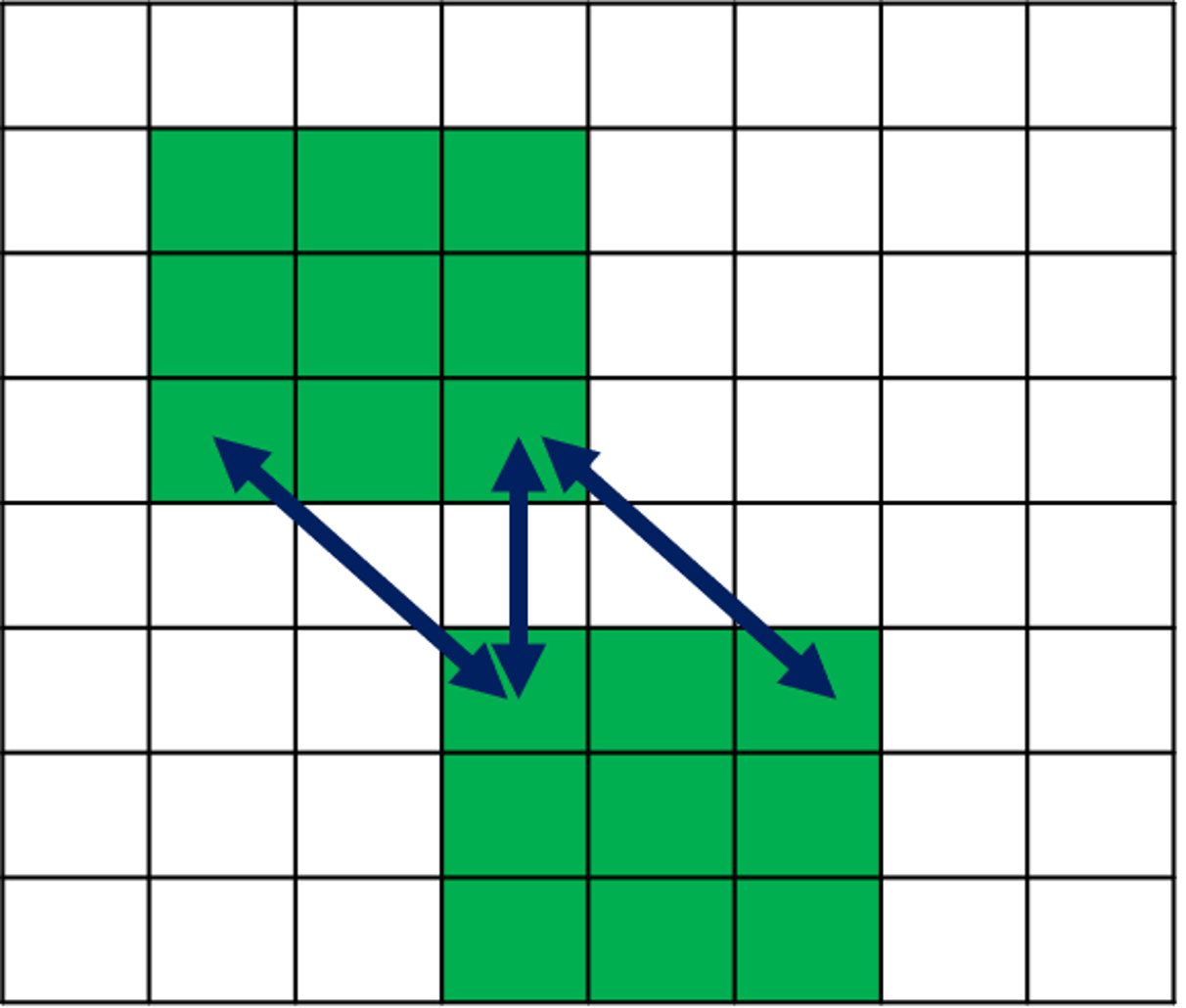}}
\hspace{10pt}
\subfloat[Self matches]{%
\label{fig:self-match}
\includegraphics[width=0.35\textwidth]{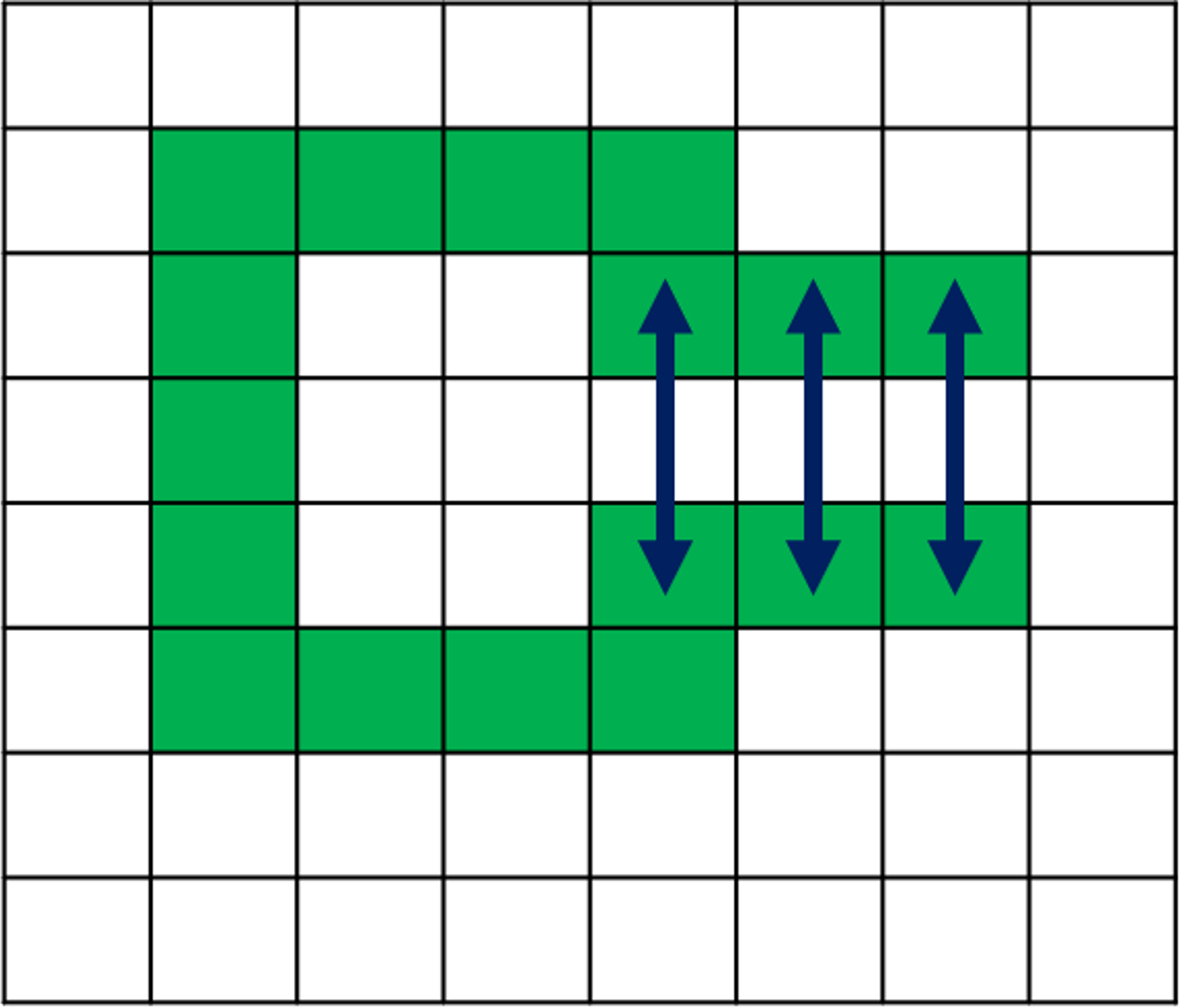}}
\caption{Obstacle matches}
\label{fig:typeMatches}
\end{figure}

At the core of the algorithm is the concept of an \textit{obstacle match}, which denotes a pair of obstacle units that potentially represent the start and end points of a narrow passage on the map. Obstacle matches are of two types: \textit{foreign matches} and \textit{self matches}. A \textit{foreign match} is a match between obstacle units from different obstacle components, while a \textit{self-match} is a match between obstacle units within the same obstacle component. Examples of these match types are shown in Figure \ref{fig:typeMatches}, with specific instances of foreign matches shown in Figure \ref{fig:foreign-match} and self-matches shown in Figure \ref{fig:self-match}. Although there are more matches in the occupancy grid maps provided, only three examples are shown. The detailed methodology for calculating these matches is explained in the following sections.

Additionally, the algorithm employs a \textit{passage value matrix}, a two-dimensional data structure equal to the dimensions of occupancy grid map. Initially, all elements of this matrix are set to the value of the shorter side of the map, resulting in a data structure that contains narrow passage lengths. 

\begin{algorithm}
\caption{Narrow Passage Detection Algorithm}\label{algo1}
\begin{algorithmic}[1]
\State foreignMatches=[ ]; selfMatches=[ ]
\State components=findConnectedComps(map,obstacle)
\State foreignMatches=foreignMatcher(components)
\For{component in components}
    \State selfMatches.append(selfMatcher(component))
\EndFor
\State passageValues=collisionCheck(foreignMatches $\cup$ selfMatches)
\State \Return passageValues
\end{algorithmic}
\end{algorithm}

\subsubsection{Finding Connected Components}\label{sec:finding-connected-components}
In this step, obstacle components consisting of obstacle units that are connected to each other are calculated via \textit{findConnectedComps()} function. There are methods developed for this purpose \citep{haralick1992computer}.

\subsubsection{Calculation of Foreign Matches}\label{sec:calculation-of-foreign-matches}
The method of determining foreign matches is that each obstacle unit is matched with the closest obstacle unit that is not in its self component. If there is only one obstacle component on the map, it means that no foreign matches can be made. The pseudocode showing the stages of the algorithm is given in Algorithm \ref{algo2}. 

\begin{algorithm}
\caption{foreignMatcher}\label{algo2}
\begin{algorithmic}[1]
\State foreignMatches=[ ]
\State borderedComponents=border(components)
\For{component in borderedComponents}
    \For{unit in component}
        \State nearest=nearestInOtherComponents(unit)
        \State dist=getDistance(unit,nearest)
        \If{dist $\leq$ maxDistParam}
            \State foreignMatches.append(\{unit,nearest\})
        \EndIf
    \EndFor
\EndFor
\State \Return foreignMatches
\end{algorithmic}
\end{algorithm}

\begin{figure}
\centering
\subfloat[Component]{%
\label{fig:borderization-1}
\includegraphics[width=0.35\textwidth]{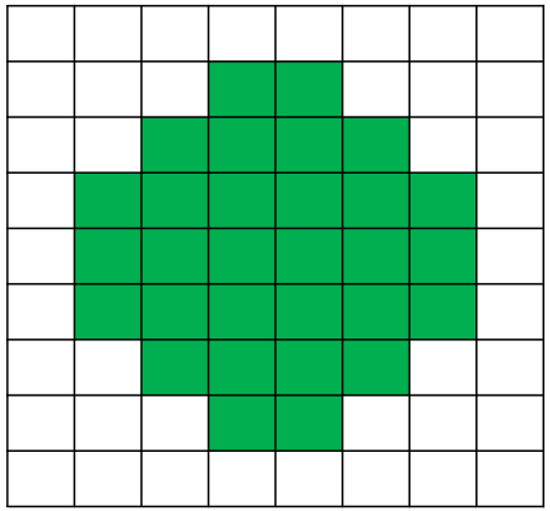}}
\hspace{10pt}
\subfloat[Bordered component]{%
\label{fig:borderization-2}
\includegraphics[width=0.35\textwidth]{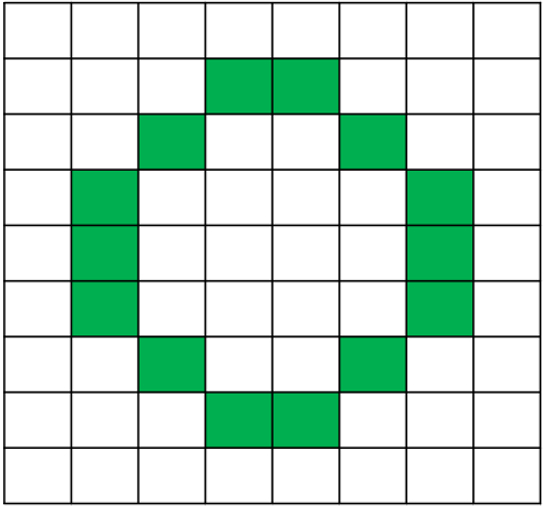}}
\caption{Obtaining boundary of the component}
\label{fig:borderization}
\end{figure}

The first step of the \textit{foreignMatcher()} function is to determine the boundaries of components adjacent to free space units. This step is done by using the function called \textit{border()}. In this way, the number of units of the components is reduced and the running time of the algorithm is considerably shortened. Reducing the number of units with the \textit{border()} function does not affect the result because it is not possible for an obstacle unit that is not adjacent to the free space to form the starting or ending point of a narrow passage. An example of this operation can be seen in Figure \ref{fig:borderization}.

The second step of the \textit{foreignMatcher()} is to find the closest obstacle unit for each unit of the bordered component, but the closest obstacle unit must be a unit of other components. \textit{nearestInOtherComponents()} function is used to select each obstacle unit closest to itself among the obstacle units belonging to other components. This function takes the obstacle unit as input whose nearest neighbor is calculated. This operation is applied for each obstacle unit in each obstacle component. These obstacle units are grouped with their matching units and added to the \textit{foreignMatches} list.

The reason for the distance control is to reduce the number of matches given to the \textit{collisionCheck()} function thus decreasing the running time of the algorithm. Additionally, with this filtering process, the matches focus on narrow passages. The filtering process is performed with \textit{maxDistanceParam}.

\subsubsection{Calculation of Self Matches}\label{sec:calculation-of-self-matches}
Complex-shaped obstacle components may have narrow passages within them. Therefore, only foreign matches are not enough to find all narrow passages in the environment. Narrow passages within the same component are calculated by using the \textit{selfMatcher()} function. The general strategy of the function is extracting parts from the component and calculating foreign matches using those parts. The pseudocode of the \textit{selfMatcher}() function is given in Algorithm \ref{algo3}.

\begin{algorithm}
\caption{selfMatcher}\label{algo3}
\begin{algorithmic}[1]
\State selfMatches = [ ]
\State cMap = createMap(component)
\State freeComps = findConnectedComps(cMap, free)
\For{freeComponent in freeComps}
    \State convex = convexHull(freeComponent)
    \State [insidePoints, outsidePoints] =  separatePoints(component, convex)
    \State insideComps =  findConnectedComps(insidePoints, obstacle)
    \State selfMatches.append(foreignMatcher(insideComps))
    \State selfMatches.append(foreignMatcher(\{insidePoints, outsidePoints\}))
    \For{comp in insideComps}
        \State selfMatches.append(selfMatcher(comp))
    \EndFor
\EndFor
\State \Return selfMatches
\end{algorithmic}
\end{algorithm}

\begin{figure}
\centering
\subfloat[Calculated convex]{%
\label{fig:extractComp-1}
\includegraphics[width=0.3\textwidth]{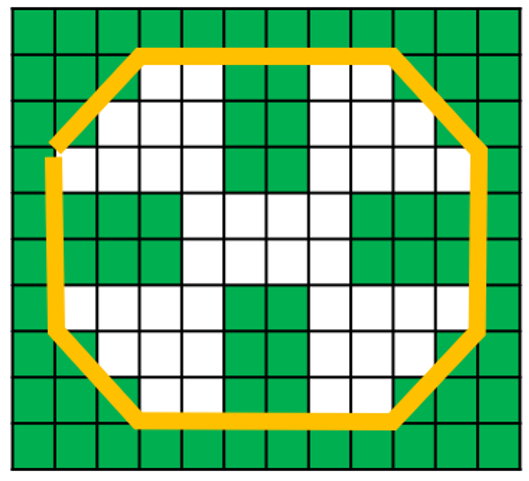}}
\hspace{10pt}
\subfloat[Separation of inside and outside units]{%
\label{fig:extractComp-2}
\includegraphics[width=0.3\textwidth]{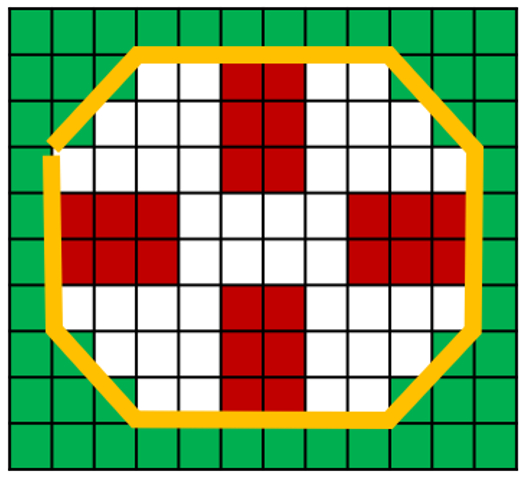}}
\hspace{10pt}
\subfloat[Extracted components]{%
\label{fig:extractComp-3}
\includegraphics[width=0.3\textwidth]{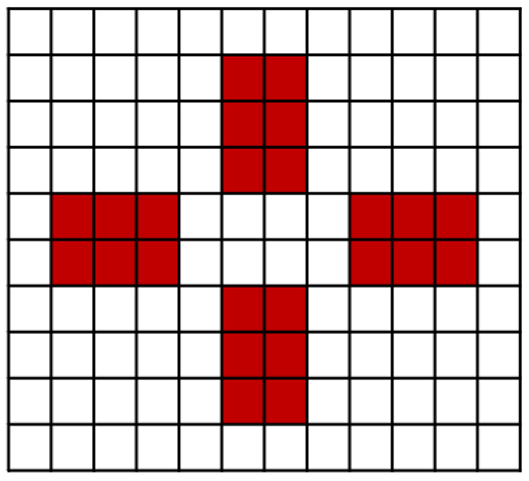}}
\caption{Process of extracting components}
\label{fig:extractComp}
\end{figure}

Initially, the smallest size map containing only the given component is created with \textit{createMap()} function. Then, free space units are clustered to get free space components in the created map. After that, the extracting process is performed for each free space component. To determine the parts to be extracted, the procedure initiates with calculating a convex polygon containing the free space component via the \textit{convexHull()} function. Subsequently, the obstacle units inside and outside the boundaries of the convex polygon are determined with \textit{separatePoints()} function. The process of extracting parts from the component is given in Figure \ref{fig:extractComp}. 

\begin{figure}
\centering
\subfloat[Calculated convex]{%
\label{fig:foreignMatchTypes-1}
\includegraphics[width=0.35\textwidth]{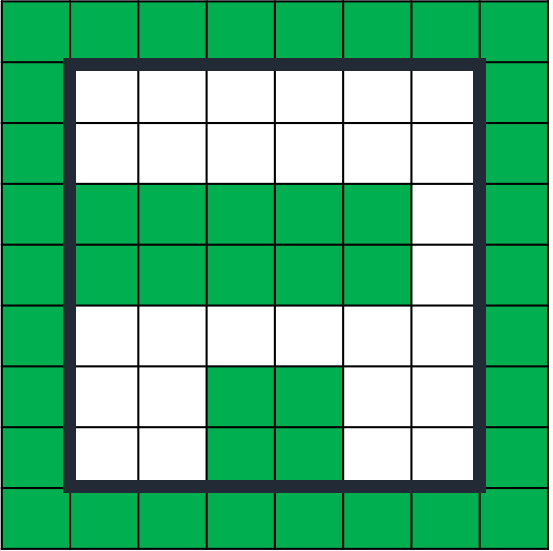}}
\hspace{10pt}
\subfloat[Types of self matches]{%
\label{fig:foreignMatchTypes-2}
\includegraphics[width=0.35\textwidth]{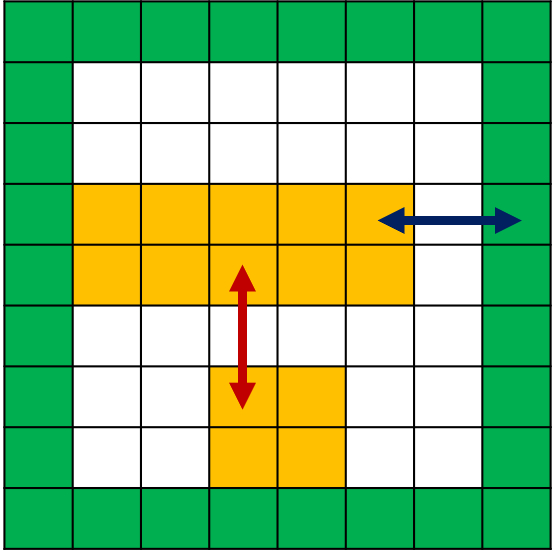}}
\caption{Foreign matcher usages in self matcher}
\label{fig:foreignMatchTypes}
\end{figure}

After the obstacle units within the polygon are determined, the \textit{foreignMatcher()} function is used twice. The first usage of the \textit{foreignMatcher()} calculates matches between the extracted parts from the component. The second usage calculates matches between inside and outside obstacle units. These two use cases are given in Figure \ref{fig:foreignMatchTypes}. An example of a match made between the extracted parts is shown in red. Additionally, an example match between the inside point and the outside point is shown in blue.

\begin{figure}
\centering
\subfloat[Extracting component]{%
\label{fig:recursiveExtractComp-1}
\includegraphics[width=0.35\textwidth]{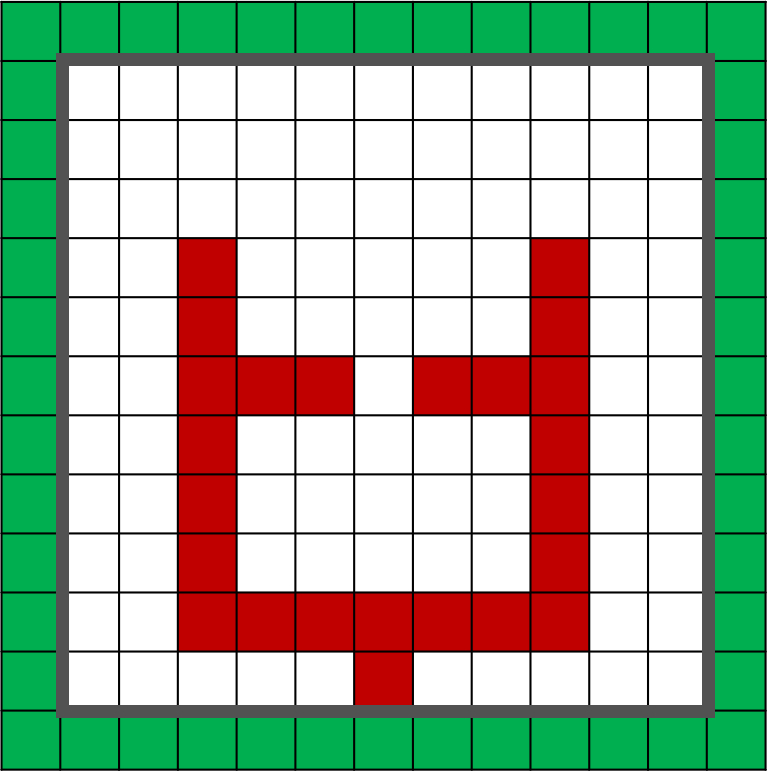}}
\hspace{10pt}
\subfloat[Extracting children of the component]{%
\label{fig:recursiveExtractComp-2}
\includegraphics[width=0.35\textwidth]{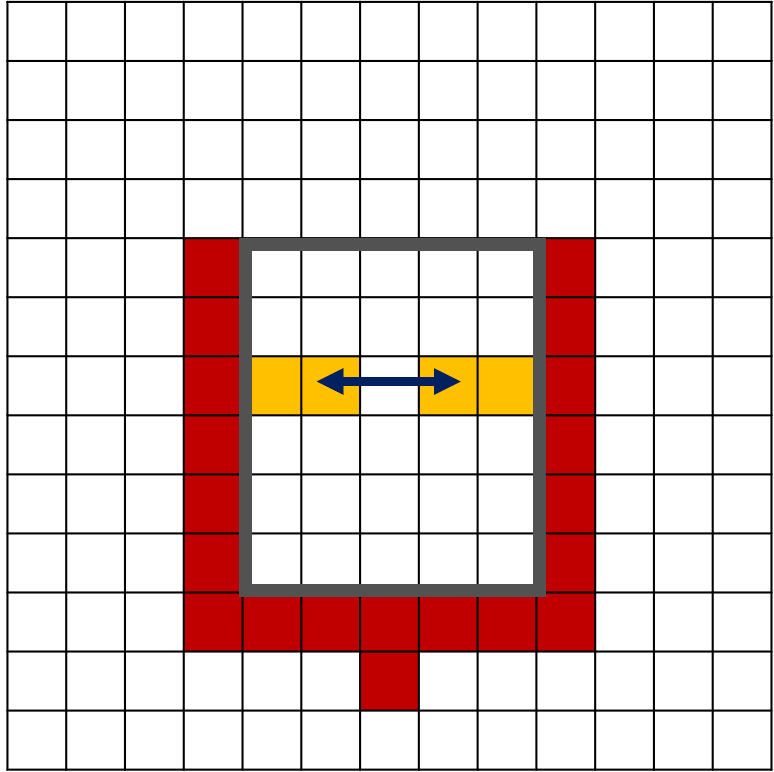}}
\caption{Recursively extracting parts from the component}
\label{fig:recursiveExtractComp}
\end{figure}

After using \textit{foreignMatcher()} function, \textit{selfMatcher()} function is called recursively for each part extracted because the part may also contain a narrow passage within itself. An example of a narrow passage inside the extracted part is given in Figure \ref{fig:recursiveExtractComp}. In summary, \textit{selfMatcher()} function is a function that finds narrow passages recursively within the component itself. The end condition of recursion is that there are no parts left to be extracted.

\subsubsection{Collision Checking of Matches}\label{sec:collision-checking-of-matches}
In this stage, \textit{collisionCheck()} function is used to inspect the cells between matched units and in case of a valid match, the passage value matrix is updated. First, a line is drawn between two obstacle units using the Bresenham Algorithm \citep{bresenham1998algorithm}. Then, the points crossing the line on the occupancy grid map are checked. If all of them are free space units, it is considered to be a narrow passage. This process is repeated for each obstacle match.

\begin{figure}
\centering
\subfloat[Matching two units in different components]{%
\label{fig:positiveCollision-1}
\includegraphics[width=0.35\textwidth]{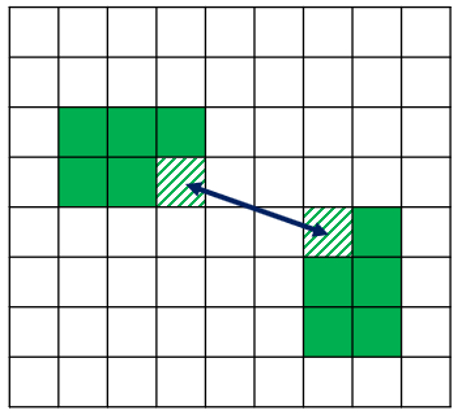}}
\hspace{10pt}
\subfloat[Inspection of the cells in between]{%
\label{fig:positiveCollision-2}
\includegraphics[width=0.35\textwidth]{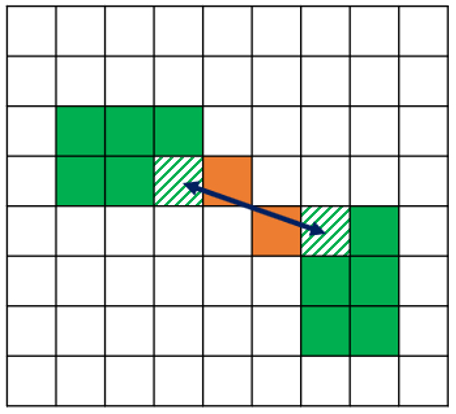}}
\caption{Valid collision check}
\label{fig:positiveCollision}
\end{figure}

\begin{figure}
\centering
\subfloat[Matching two units in different components]{%
\label{fig:negativeCollision-1}
\includegraphics[width=0.35\textwidth]{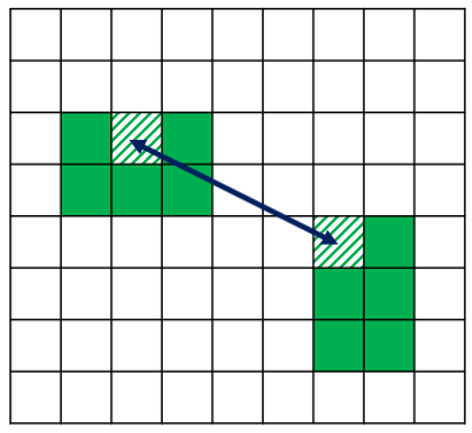}}
\hspace{10pt}
\subfloat[Inspection of the cells in between]{%
\label{fig:negativeCollision-2}
\includegraphics[width=0.35\textwidth]{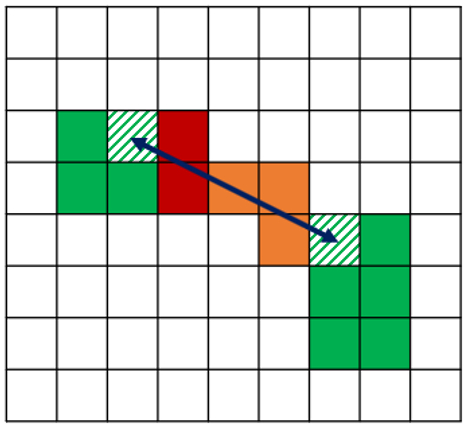}}
\caption{Invalid collision check}
\label{fig:negativeCollision}
\end{figure}

A foreign match is given in Figure \ref{fig:positiveCollision-1}. In Figure \ref{fig:positiveCollision-2}, the units between this match are determined and it can be seen that they are all on the free space. The given example in Figure \ref{fig:positiveCollision} is a valid match. Figure \ref{fig:negativeCollision} shows an example of an invalid match. The test returns collision because there are obstacle units shown in red between the matches. Therefore, this match is not considered a narrow passage.

\begin{figure}
\centering
\subfloat[Calculation of the distance between two units]{%
\label{fig:distance-1}
\includegraphics[width=0.35\textwidth]{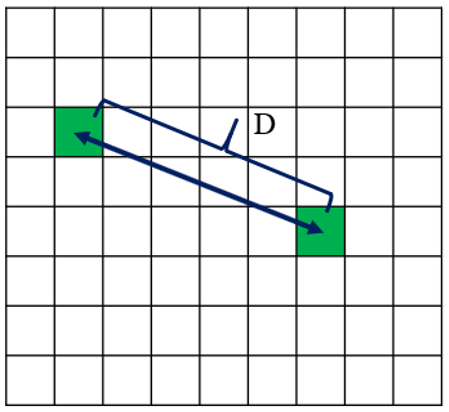}}
\hspace{10pt}
\subfloat[Replacing the values in the passage value matrix]{%
\label{fig:distance-2}
\includegraphics[width=0.35\textwidth]{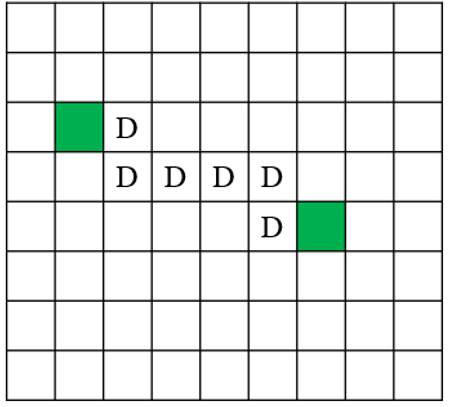}}
\caption{Updating passage value matrix}
\label{fig:distance}
\end{figure}

\begin{figure}
\centering
\subfloat[The map to identify narrow passages]{%
\label{fig:passage-value-matrix-1}
\includegraphics[width=0.35\textwidth]{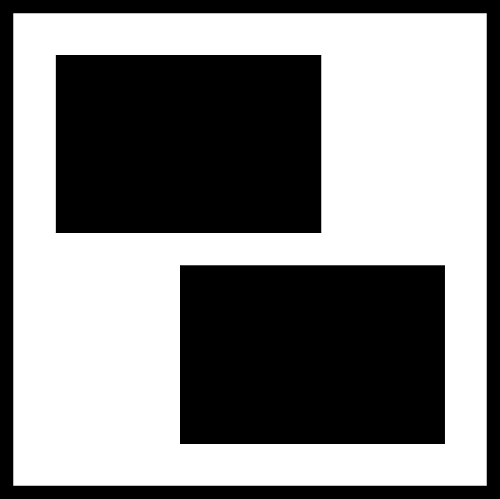}}
\hspace{10pt}
\subfloat[Visualization of the passage value matrix]{%
\label{fig:passage-value-matrix-2}
\includegraphics[width=0.465\textwidth]{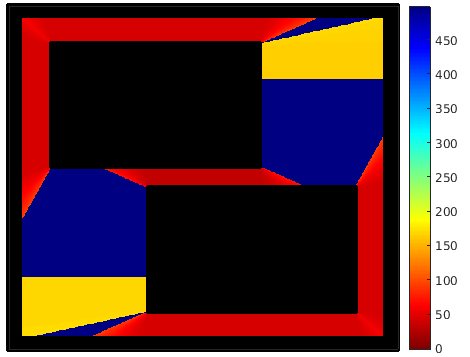}}
\caption{Detection of narrow passages on the map and obtaining the passage value matrix}
\label{fig:passage-value-matrix}
\end{figure}

In cases where the \textit{collisionCheck()} is successful, the passage value matrix needs to be updated. The passage value matrix is a matrix that is the same size as the occupancy grid map. In this matrix, given in Figure \ref{fig:distance}, the distance (D) value between the matches is placed on all coordinates on the line drawn between the matches. The passage value matrix obtained by calculating all matches on a sample map is shown in Figure \ref{fig:passage-value-matrix}. 

There are values already placed in these units. Consequently, the previous distance value is compared with the current distance value. The smaller value is valid since the narrowest passage containing the relevant coordinate is sought to be determined.

\subsection{Sampling Strategy}\label{sec:sampling-strategy}
The method presented is based on the detection of narrow passages and guided sampling of these regions. The detection of the narrow passages is explained in Section \ref{sec:passage-identification-algorithm}. The sampling strategy is explained by using the narrow passage value matrix.

Initially, the passage value matrix $A$ is normalized so that the sum of its elements is equal to one. The result is the two-dimensional discrete probability density function $P$. This process is given in Equation \ref{eq:normalization}. Here, the $m \times n$ dimensional matrix $A$ represents the passage value matrix where $a_{ij}$ is the $i^{th}$ row $j^{th}$ column element of $A$. Also, the value $S$ is the sum of the elements in matrix $A$. Each element $a_{ij}$ in $A$ is divided by $S$ to find each element $p_{ij}$ in the probability density function $P$.

\begin{align}
\begin{split}
S &= \sum_{i=1}^{m} \sum_{j=1}^{n} a_{ij} \\
p_{ij} &= \frac{a_{ij}}{S}, \quad \text{for } i = 1, \ldots, m \text{ and } j = 1, \ldots, n
\end{split}
\label{eq:normalization}
\end{align}

Subsequently, weighted sampling is performed using the density function $P$ calculated in Equation \ref{eq:normalization}. For this purpose, the cumulative distribution $C$ is obtained first, given in Equation \ref{eq:cumulative}. 

\begin{align}
\begin{split}
c_{k} &= \sum_{x=1}^{i-1} \sum_{y=1}^{n} p_{xy} + \sum_{y=1}^{j} p_{iy},\\ \text{ for } k &= 1, \ldots, {m \times n} \\
\text{where } i &= \left\lfloor \frac{k}{n} \right\rfloor \text{ and } j = k \bmod n\\
\end{split}
\label{eq:cumulative}
\end{align}

In the $C = \{c_1, c_2, \ldots, c_n\}$ cumulative distribution array, $c_k$ represents the $k^{th}$ element value of the array and $c_n = 1$. To determine the index of the cell being sampled, a random variable $U$ between 0 and 1 is selected. Then, the first value $c_k$ greater than $U$  is found as in Equation \ref{eq:index-selecting}. Here, the value $k$  represents the index to be sampled. 

\begin{align}
 k_{selected} = \min \{ k : U \leq c_k, k = 1, 2, \ldots, n \}
\label{eq:index-selecting}
\end{align}

The process of converting the index selected from the cumulative distribution back to the 2D position on the map is given in Equation \ref{eq:index-converting}.

\begin{align}
\begin{split}
i = \left\lfloor \frac{k_{selected}}{n} \right\rfloor\text{, and } j = k_{selected} \bmod n
\end{split}
\label{eq:index-converting}
\end{align}

Narrow passage sampling is performed as described in the given equations using the passage value matrix. However, narrow passage sampling alone is not sufficient to cover the map for PRM. Similarly, uniform sampling alone is not sufficient. Therefore, a hybrid sampling method is preferred to combine strengths of both. 

To determine the ratio between uniform and narrow passage sampling, a series of simulations is conducted in which different proportions were investigated. In Table \ref{tab:hybrid-sampling-ratio}, results on varying hybrid sampling ratios are provided. It is observed that each map exhibited a distinct optimal proportion. The hybrid sampling technique is implemented at a rate of 1:1 to make it simple. In this way, with the uniform sampler, the roadmap is spread more, and with the narrow passage sampler, critical regions are sampled and subgraphs are connected.

MBPI works on the idea that narrow passages can be identified through occupancy grid map during discretization. The ability to detect these passages depends on the grid's resolution. If the resolution is very high, narrow passages can be detected and prioritized during sampling. However, a very high resolution increases computational cost, prolonging the identification process. Conversely, if the resolution is too low to capture the details of narrow regions, the detection algorithm cannot identify them, causing the narrow passage sampler to behave like a uniform sampler. In this case, the hybrid sampling strategy simply becomes a uniform sampling strategy.

\section{Experiments}\label{sec:experiments}
Three types of experiments are presented. Firstly, the benchmark environment to compare MBPI with the existing methods is introduced. Then, the results obtained in simulation and real-world test scenarios are given. 

\subsection{Benchmark Environment}\label{sec:benchmark-environment}
The Open Motion Planning Library (OMPL) \citep{sucan2012open} is an open-source library that contains many sampling-based algorithms. This library provides a platform that allows comparing the performance of algorithms. MBPI is integrated by adding a new sampler to the OMPL library to compare with other methods \citep{git-repo}. Within the scope of the experiments, the performance of the algorithm is compared with Bridge Test, Gaussian, Obstacle-based, and Uniform samplers within OMPL. The parameters of these samplers are kept constant at their default values in OMPL. During this evaluation process, PRM is used with default parameters in OMPL. \textit{maxDistanceParam}, which determines the maximum narrow passage width of the detection algorithm, is kept constant at $0.05$ (i.e. $5\%$ of the map size) in all experiments. Performance analysis is carried out through the metrics of planning time, number of milestones in the roadmap, and path length which is obtained by the solution length parameter of the OMPL. The system and software configurations used in the experiments are summarized in Table \ref{tab:system-configuration}.

\begin{table}[ht]
    \centering
    \caption{System and Software Configurations Used in Experiments}
    \label{tab:system-configuration}
    \begin{tabular}{l l}
    \hline
    \textbf{Component} & \textbf{Detail} \\
    \hline
    RAM & 64GB \\
    Processor & Intel Core i7-10700F CPU @ \\
    & 2.90GHz x 16 \\
    Graphics Card & NVIDIA GeForce GT 610 \\
    Operating System & Ubuntu 20.04.6 LTS (64-bit) \\
    OMPL Version & 1.6.0 \\
    ROS Version & ROS Noetic \\
    Robot Model & Turtlebot3 Burger \\
    \hline
    \end{tabular}
\end{table}

\begin{figure*}
\centering
\subfloat[Env.~1]{%
\label{fig:specific-03}
\includegraphics[width=0.32\textwidth]{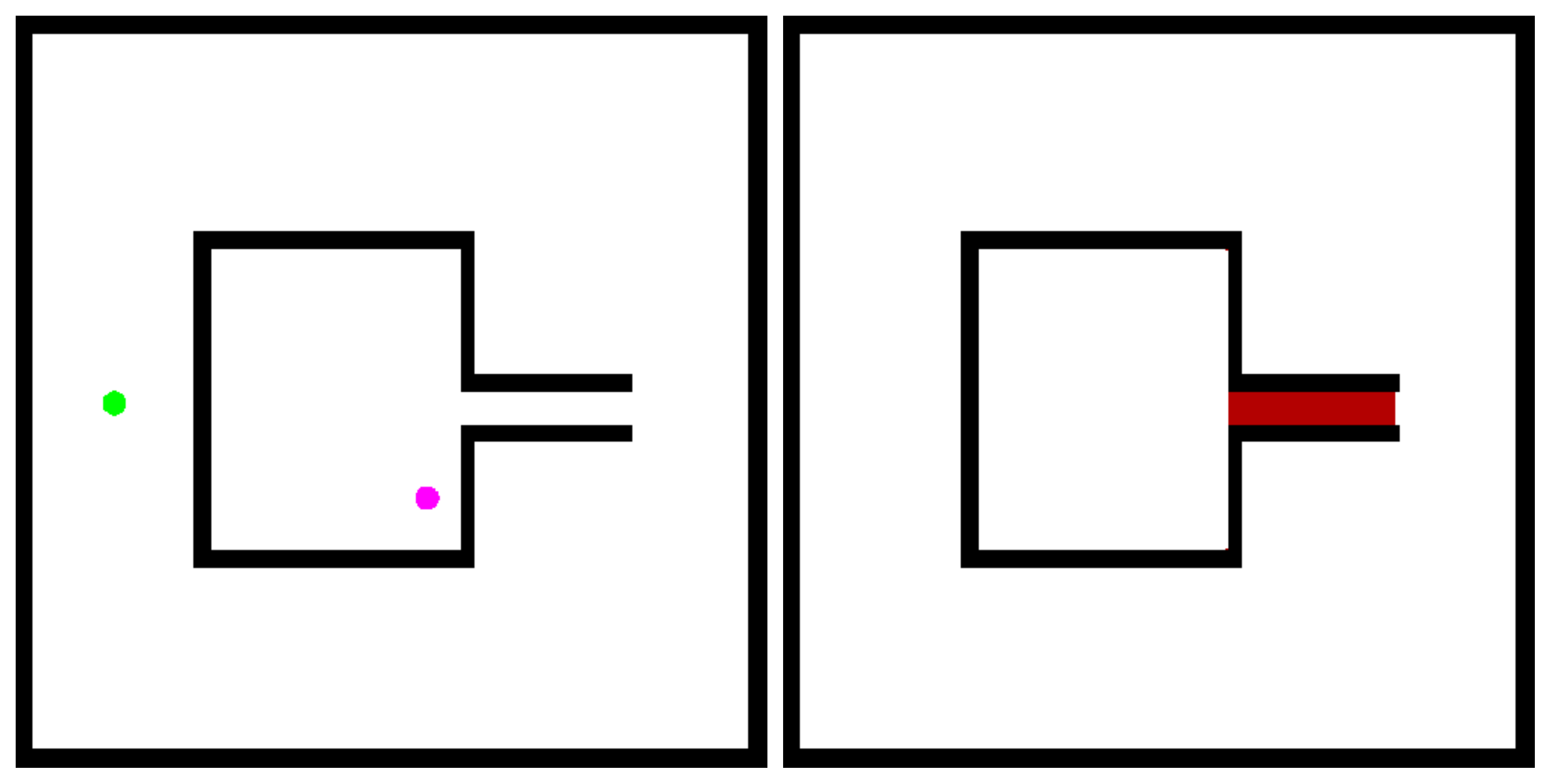}}\hfill
\subfloat[Env.~2]{%
\label{fig:specific-04}
\includegraphics[width=0.32\textwidth]{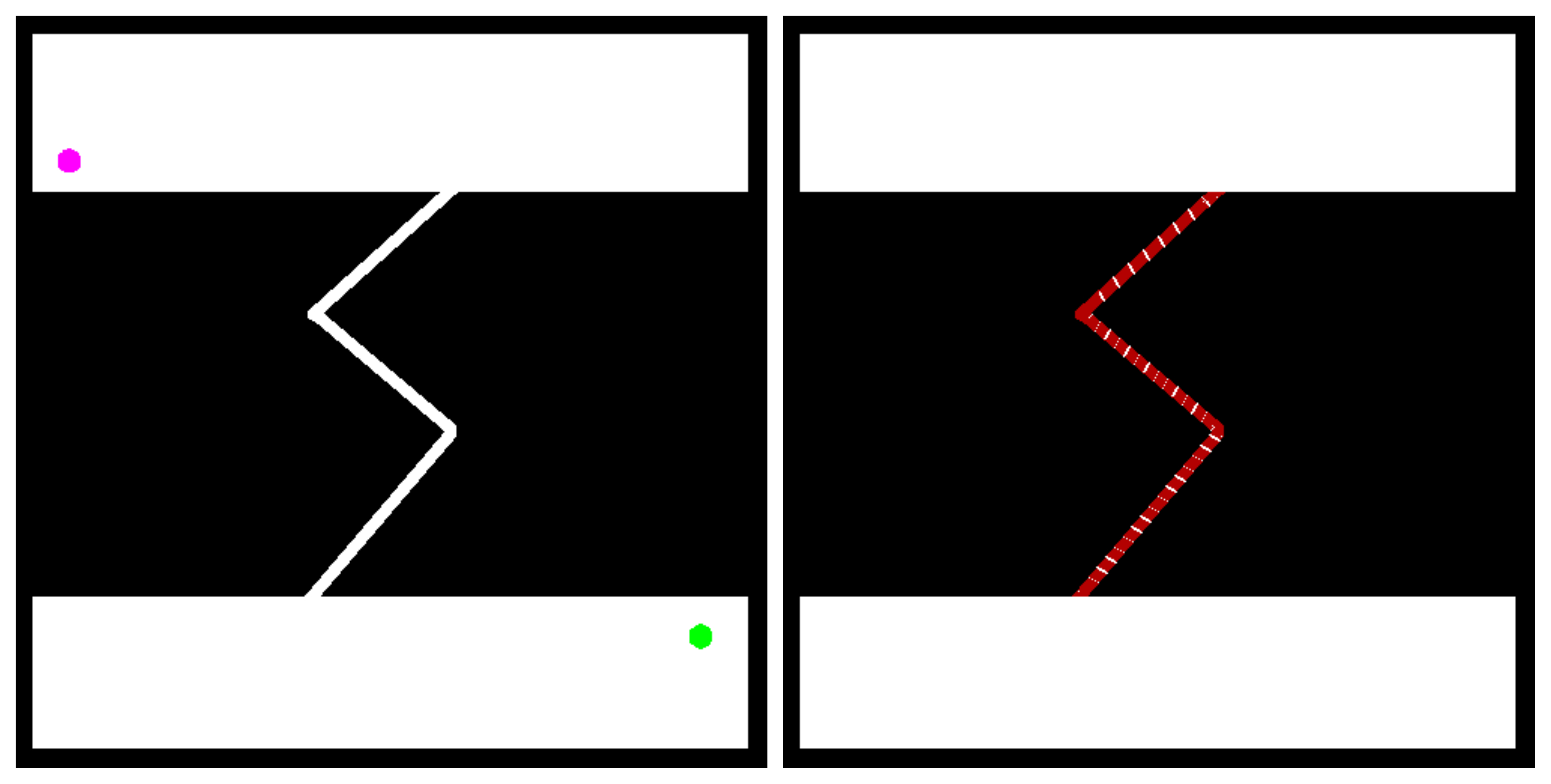}}\hfill
\subfloat[Env.~3]{%
\label{fig:specific-05}
\includegraphics[width=0.32\textwidth]{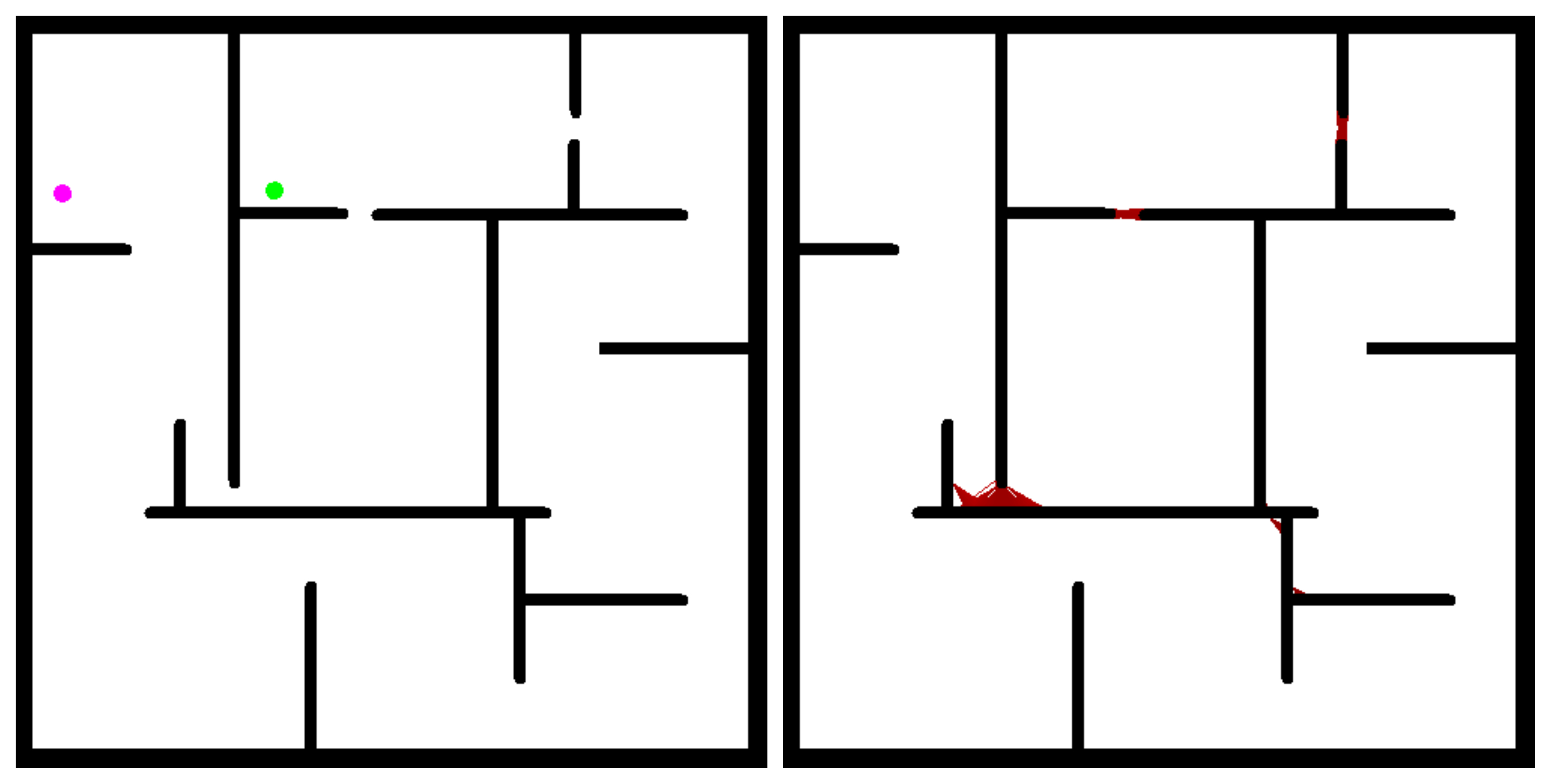}}\\[5pt]
\subfloat[Env.~4]{%
\label{fig:specific-010}
\includegraphics[width=0.32\textwidth]{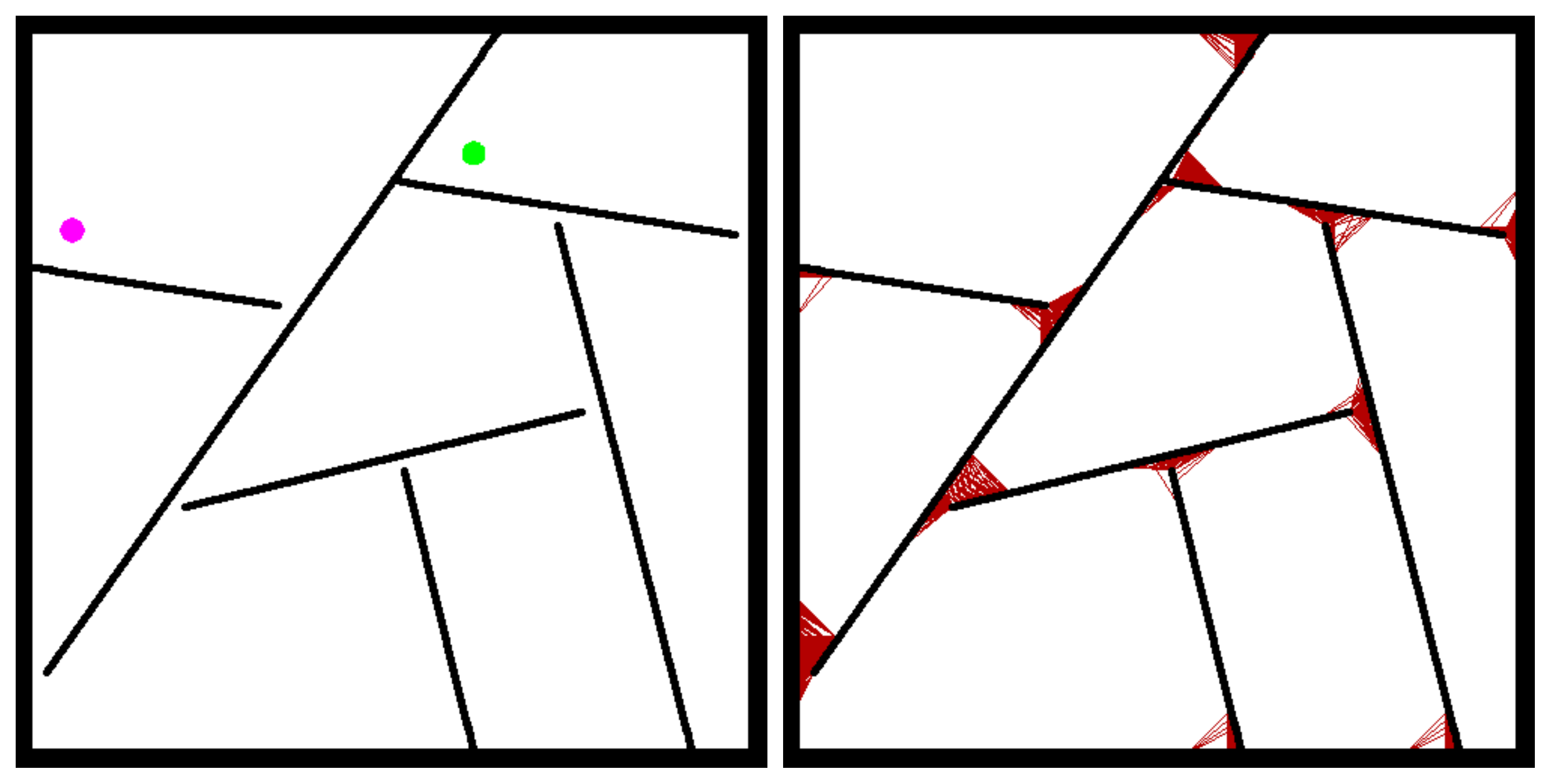}}\hfill
\subfloat[Env.~5]{%
\label{fig:specific-02}
\includegraphics[width=0.32\textwidth]{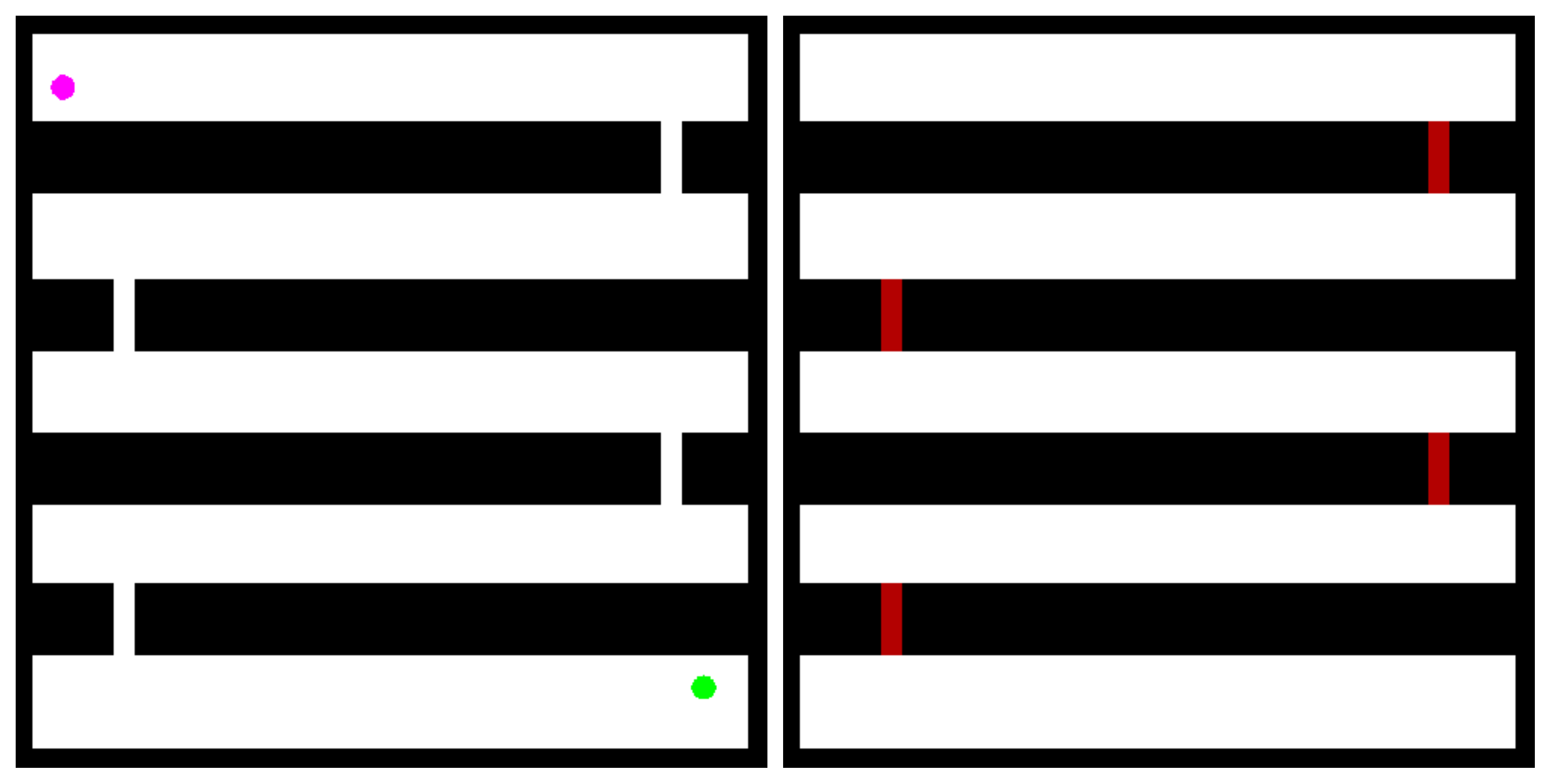}}\hfill
\subfloat[Env.~6]{%
\label{fig:specific-r3}
\includegraphics[width=0.32\textwidth]{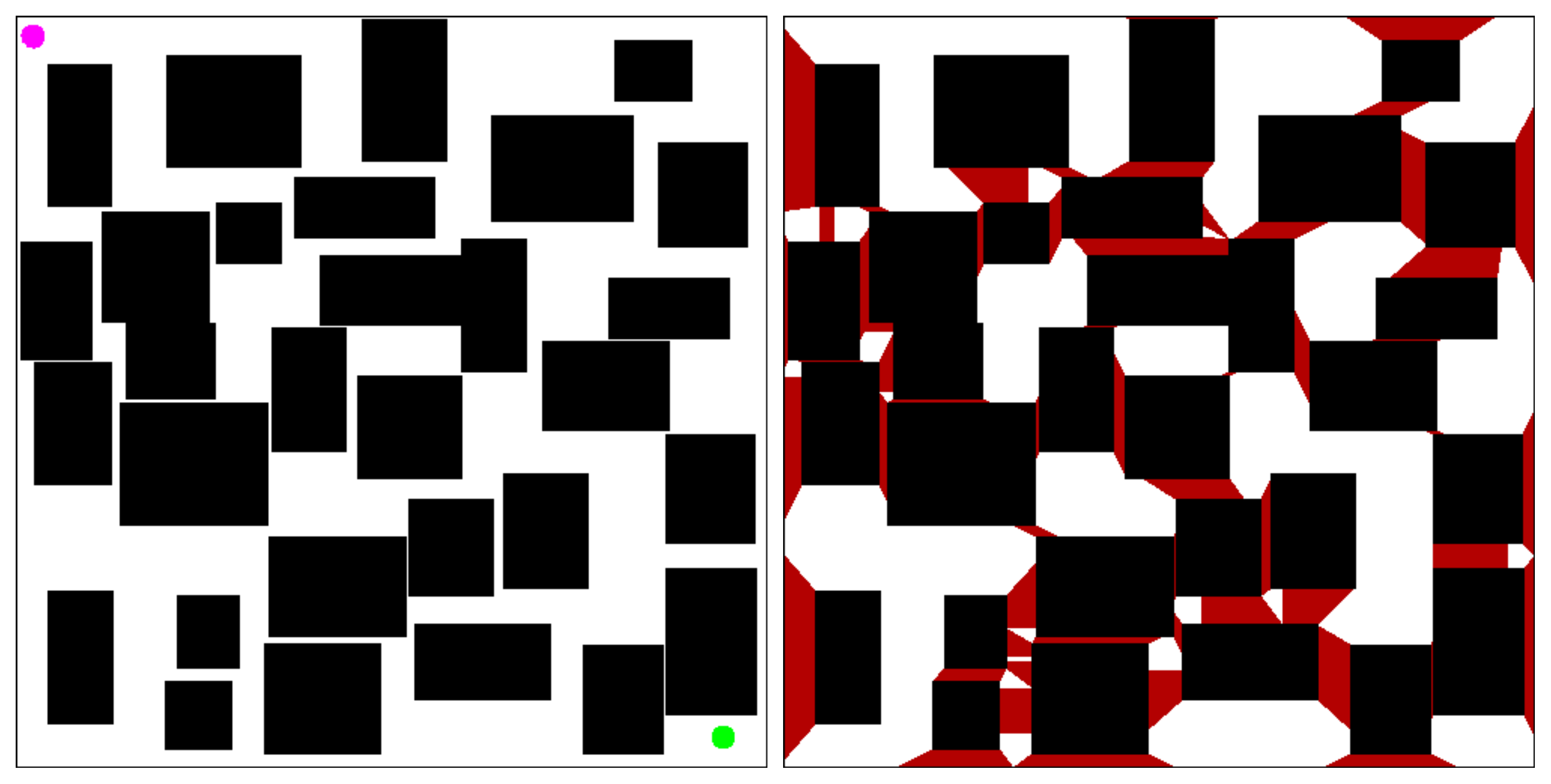}}
\caption{The environments used in the experiment (left) and the detected narrow passage zones by MBPI, marked in red (right). The initial and target points in each environment are indicated by magenta and green markers respectively.}
\label{fig:specific-maps}
\end{figure*}

\subsection{Test Scenarios and Results}\label{sec:test-scenarios-and-results}
The performances of MBPI and the existing methods are compared in 3 different types of experiments. 

\begin{table*}[htbp]
    \centering
    \caption{Performances of each sampler on each environment. Narrow passage detection time of MBPI is given in parenthesis for each environment in the sampler column.}
    \begin{tabular}{|c|c|ccc|}
    \hline
    \multirow{2}{*}{\textbf{Env. (Fig.)}} & \multirow{2}{*}{\textbf{Sampler}} & \multicolumn{3}{c|}{\textbf{Metrics} (Avg.(Std. dev.))} \\ 
    & & Planning Time (s) & Number of Milestones & Path Length (m) \\
    \hline
    \multirow{5}{*}{1} & MBPI $(0.126s)$ & \(0.0060 (0.0022)\) & \(224.15 (95.01)\) & \(18.77 (1.59)\) \\
    & Bridge Test & \(0.0091 (0.0145)\) & \(205.97 (518.98)\) & \(22.40 (3.24)\) \\
    & OBPRM & \(0.0333 (0.0262)\) & \(1604.50 (1483.21)\) & \(21.80 (5.31)\) \\
    & Gaussian & \(0.0140 (0.0116)\) & \(529.38 (639.58)\) & \(18.54 (2.43)\) \\
    & Uniform & \(0.0127 (0.0084)\) & \(439.77 (434.16)\) & \(17.67 (1.59)\) \\
    \hline
    \multirow{5}{*}{2} & MBPI $(0.264s)$ & \(0.0046 (0.0015)\) & \(160.08 (68.86)\) & \(24.18 (1.74)\) \\
        & Bridge Test & \(0.0136 (0.0290)\) & \(511.31 (1518.74)\) & \(25.82 (2.02)\) \\
    & OBPRM & \(0.0216 (0.0073)\) & \(1197.45 (537.76)\) & \(22.53 (0.74)\) \\
    & Gaussian & \(0.0197 (0.0076)\) & \(939.31 (559.35)\) & \(22.96 (1.02)\) \\
    & Uniform & \(0.0299 (0.0134)\) & \(1731.24 (1108.67)\) & \(22.67 (0.78)\) \\
    \hline
    \multirow{5}{*}{3} & MBPI $(0.236s)$ & \(0.0054 (0.0022)\) & \(239.58 (125.72)\) & \(15.30 (4.58)\) \\
    & Bridge & \(0.0102 (0.0060)\) & \(279.77 (215.85)\) & \(33.31 (9.96)\) \\
    & OBPRM & \(0.0128 (0.0102)\) & \(599.55 (613.03)\) & \(37.21 (15.38)\) \\
    & Gaussian & \(0.0057 (0.0032)\) & \(172.53 (133.42)\) & \(36.00 (6.50)\) \\
    & Uniform & \(0.0059 (0.0025)\) & \(192.18 (100.58)\) & \(32.48 (8.63)\) \\
    \hline
    \multirow{5}{*}{4} & MBPI $(0.308s)$ & $0.0072 (0.0028)$ & $306.95 (159.18)$ & $34.68 (2.01)$ \\
    & Bridge PRM & $0.1059 (0.0861)$ & $4168.47 (3542.68)$ & $34.17 (2.39)$ \\
    & OBPRM & $0.0203 (0.0155)$ & $1130.24 (1042.75)$ & $38.22 (3.37)$ \\
    & Gaussian PRM & $0.0667 (0.0485)$ & $4123.15 (3503.4 )$ & $33.13 (1.64)$ \\
    & Uniform PRM & $0.0541 (0.0384)$ & $3195.84 (2732.78)$ & $33.41 (1.75)$ \\
    \hline
    \multirow{5}{*}{5} & MBPI $(0.254s)$ & \(0.0059 (0.0024)\) & \(312.22 (137.32)\) & \(53.75 (1.17)\) \\
    & Bridge Test & \(0.0111 (0.0061)\) & \(466.72 (354.17)\) & \(54.94 (1.26)\) \\
    & OBPRM & \(0.0331 (0.0224)\) & \(1932.30 (1642.11)\) & \(55.99 (2.98)\) \\
    & Gaussian & \(0.0128 (0.0082)\) & \(565.74 (492.67)\) & \(54.48 (1.60)\) \\
    & Uniform & \(0.0139 (0.0078)\) & \(618.66 (467.28)\) & \(54.18 (1.52)\) \\
    \hline
    \multirow{5}{*}{6} & MBPI $(0.387s)$ & $0.0057 (0.0021)$ & $316.45 (119.96)$ & $24.29 (2.95)$ \\
    & Bridge Test & $0.0094 (0.0051)$ & $499.52 (314.16)$ & $24.13 (3.32)$ \\
    & OBPRM & $0.0261 (0.0289)$ & $1846.79 (2359.44)$ & $23.05 (2.485)$ \\
    & Gaussian & $0.0076 (0.0043)$ & $393.83 (257.52)$ & $25.30 (3.68)$ \\
    & Uniform & $0.0079 (0.0041)$ & $413.68 (247.24)$ & $25.35 (3.81)$ \\
    \hline    
    \end{tabular}
    \label{tab:specific-results}
\end{table*}

\begin{table*}
    \centering
    \caption{Number of Milestones Under Varying MBPI Hybrid Sampling Ratios (Uniform:Narrow)}
    \renewcommand{\arraystretch}{1.3}
    \begin{tabular}{|c|ccccc|}
         \hline
         \multirow{2}{*}{\textbf{Env.}} & \multicolumn{5}{c|}{\textbf{Number of Milestones} (Avg. (Std. dev.))} \\
         & 5:1 & 3:1 & 1:1 & 1:3 & 1:5 \\
         \hline
         1 & $\mathbf{73.12 (41.55)}$ & $80.27 (57.82)$ & $79.67 (47.19)$ & $144.56 (91.96)$ & $212.67 (141.31)$ \\
         \hline
         2 & $254.93 (129.56)$ & $188.63 (104.32)$ & $114.14 (54.52)$ & $\mathbf{93.62 (48.61)}$ & $129.30 (97.03)$ \\
         \hline
         3 & $81.38 (55.34)$ & $\mathbf{71.53 (43.05)}$ & $108.09 (139.10)$ & $280.63 (413.6)$ & $374.86 (459.91)$ \\
         \hline
         4 & $322.64 (166.63)$ & $257.28 (122.68)$ & $\mathbf{211.17 (173.06)}$ & $257.17 (217.49)$ & $510.52 (646.80)$ \\
         \hline
         5 & $179.84 (76.34)$ & $\mathbf{173.01 (66.79)}$ & $238.51 (102.37)$ & $474.46 (264.27)$ & $686.21 (276.88)$ \\
         \hline
         6 & $320.51 (151.13)$ & $285.15 (124.50)$ & $\mathbf{249.90 (122.68)}$ & $294.15 (166.29)$ & $316.49 (180.52)$ \\
         \hline
    \end{tabular}
    \label{tab:hybrid-sampling-ratio}
\end{table*}

\subsubsection{Specific Map Experiments}\label{sec:specific-map-experiments}
MBPI and existing methods are compared for specific narrow passage scenarios. All of the prepared maps are $500 \times 500$ and are given in Figure \ref{fig:specific-maps}. In addition, the regions detected as narrow passages are labeled. The experiments are repeated $500$ times for the selected initial and target points in each environment and the statistics of these experiments are given in Table \ref{tab:specific-results}. Additionally, the milestones generated by the samplers in one experiment for each environment are given in Figure \ref{fig:specific-map-samples}. The results obtained for each map can be interpreted as follows:

\begin{figure*}
\centering
\subfloat[Env.~1]{%
\label{fig:env1-samples}
\includegraphics[width=0.99\textwidth]{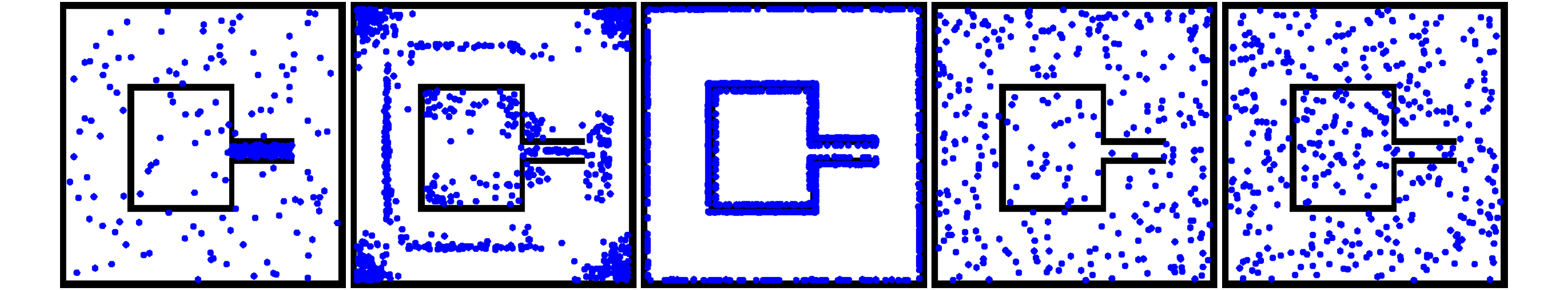}}\\[5pt]
\subfloat[Env.~2]{%
\label{fig:env2-samples}
\includegraphics[width=0.99\textwidth]{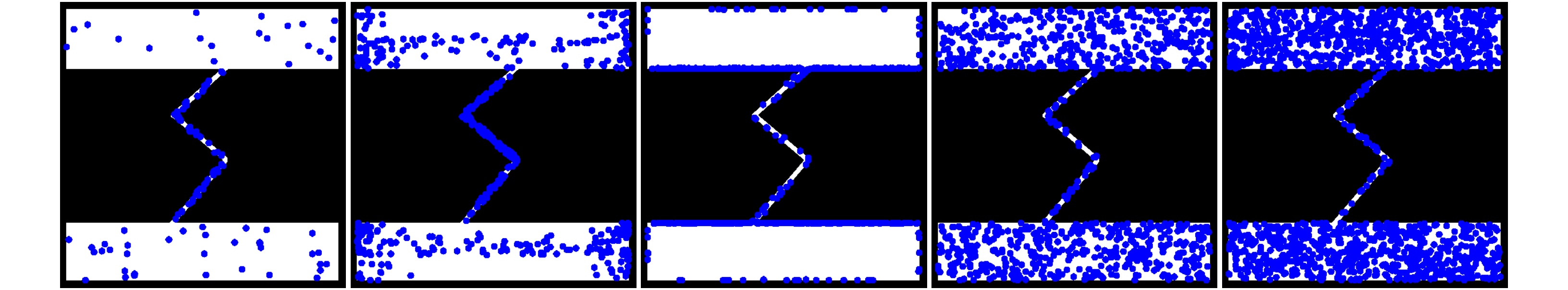}}\\[5pt]
\subfloat[Env.~3]{%
\label{fig:env3-samples}
\includegraphics[width=0.99\textwidth]{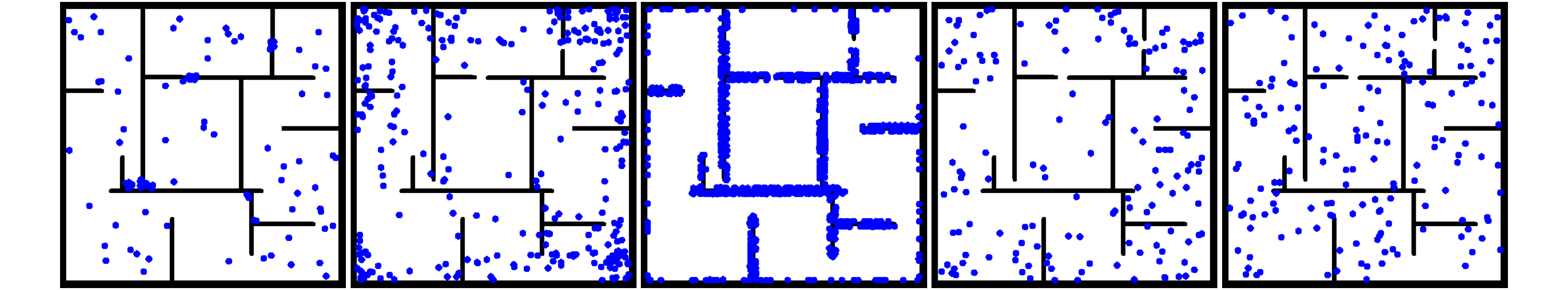}}\\[5pt]
\subfloat[Env.~4]{%
\label{fig:env4-samples}
\includegraphics[width=0.99\textwidth]{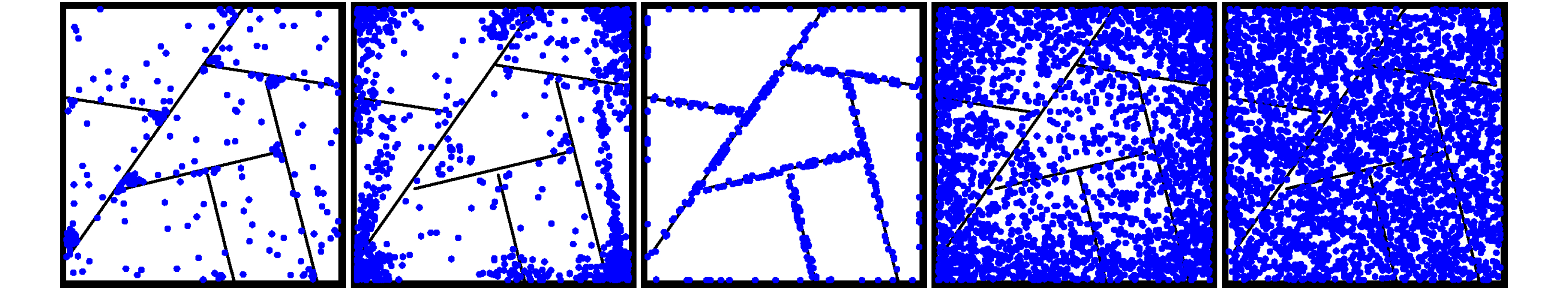}}\\[5pt]
\subfloat[Env.~5]{%
\label{fig:env5-samples}
\includegraphics[width=0.99\textwidth]{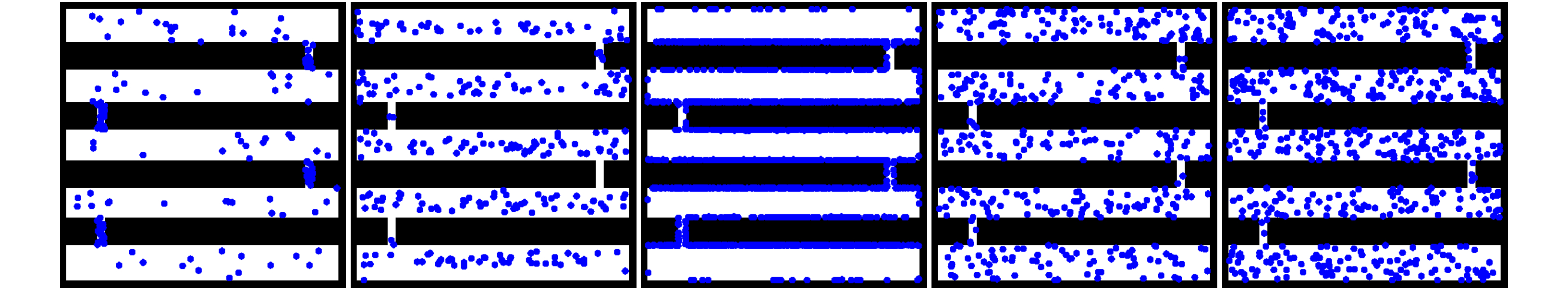}}\\[5pt]
\subfloat[Env.~6]{%
\label{fig:env6-samples}
\includegraphics[width=0.99\textwidth]{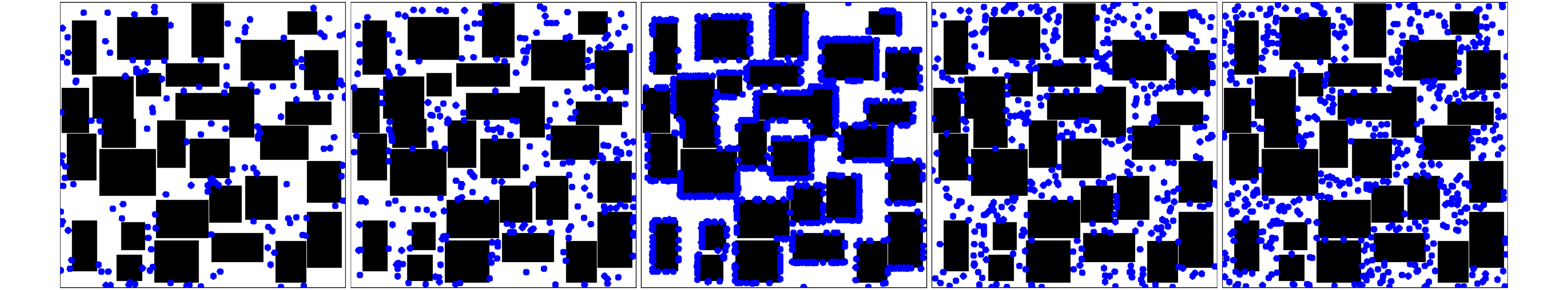}}
\caption{Milestones taken by MBPI, Bridge Test, OBPRM, Gaussian and Uniform samplers respectively until finding a valid path between the starting and ending points in each environment.}
\label{fig:specific-map-samples}
\end{figure*}

\begin{itemize}
    \item Environment 1 is a bottle-like structure with a narrow gate. In this map, MBPI successfully detected the critical region that may affect the connectivity problem as shown in Figure \ref{fig:specific-03}. Thus, narrow passage samples directly affect the performance of the algorithm positively. As seen in Table \ref{tab:specific-results}, MBPI gives the best result in planning time metric when compared with other methods. The samples generated by the methods are given in Figure \ref{fig:env1-samples}. Here, Bridge Test method generated samples in dead-end regions besides the critical narrow region. OBPRM and Gaussian, also can be seen in Figure 5, sample obstacle boundaries instead of narrow regions. This problem of Gaussian and OBPRM is common to all environments. For these reasons, our method outperforms the other methods. 

    \item Environment 2 is a narrow zigzag corridor map. In this map, MBPI successfully detected the long thin corridor, which is the critical region that may affect the connectivity problem in the map, as shown in Figure \ref{fig:specific-04}. As seen in Table \ref{tab:specific-results}, MBPI gives the best results in planning time and number of milestones metrics when compared with other methods. In this environment, Bridge Test's performance underperformed MBPI. The reason for this, the bridge test method creates a milestone when two samples coincide with the obstacle and the middle of these two samples coincides with the free space. However, there is a very thin zigzag corridor, so the probability of the middle of the two samples falling on the free space is quite low and the working performance of the bridge test is adversely affected. 

    \item Environment 3 is a map containing rooms of different sizes. As shown in Figure \ref{fig:specific-05}, MBPI successfully detects narrow passages. These narrow passages offer an alternative route from the initial to the target point. As seen in Table \ref{tab:specific-results}, MBPI is outperforming the other methods in path length metric. Because the alternative route that the other methods cannot detect is considerably shorter. The other methods reach the destination through the longer route because they cannot perform sampling focused on these narrow passages as can be seen in Figure \ref{fig:env3-samples}. Additionally, although MBPI is worse in the milestone metric, it is more efficient than the other methods in the planning time metric.

    \item Environment 4 is a map with long thin sticks placed at different angles. Figure \ref{fig:specific-010} shows the regions that MBPI detects as narrow. Taking these regions into account, hybrid sampling has provided a advantage over other methods in all metrics. The noticeable result is that the bridge test method gives much worse results in planning time and number of milestones metrics compared to other methods. To sample in the narrow passage region, the bridge test requires two samples over obstacles. In this map, the bars creating the narrow passage are very thin, hence it is challenging for the bridge test to generate two obstacle samples. However, MBPI identifies critical regions and focuses directly on those regions as can be seen in Figure \ref{fig:env4-samples}, which makes it advantageous in all metrics.

    \item Environment 5 is a map representing a simple maze. In this map, MBPI succeeded in detecting the critical regions that may influence the connectivity problem, as shown in Figure \ref{fig:specific-03}. Thus, narrow passage samples directly improve the performance of the algorithm. As can be seen in Table \ref{tab:specific-results}, MBPI gives the best result in all metrics compared to other methods.

    \item Environment 6 consists of a large number of rectangular obstacles. Figure \ref{fig:specific-r3} shows the regions detected as narrow by MBPI. Due to the large number of obstacle components, there are many narrow passages on the map. These narrow passages provide numerous alternative routes from the initial to the target points. As can be seen in Table \ref{tab:specific-results}, MBPI outperforms the other methods in planning time and number of milestone metrics.
\end{itemize}

\subsubsection{Random Map Experiments}\label{sec:random-map-experiments}
This section evaluates the performance of the algorithm on randomly generated maps through a Monte Carlo experiment. This experiment is crucial to understanding how the algorithm behaves in various and unpredictable environments. 

\begin{figure}
\centering
\includegraphics[width=11cm]{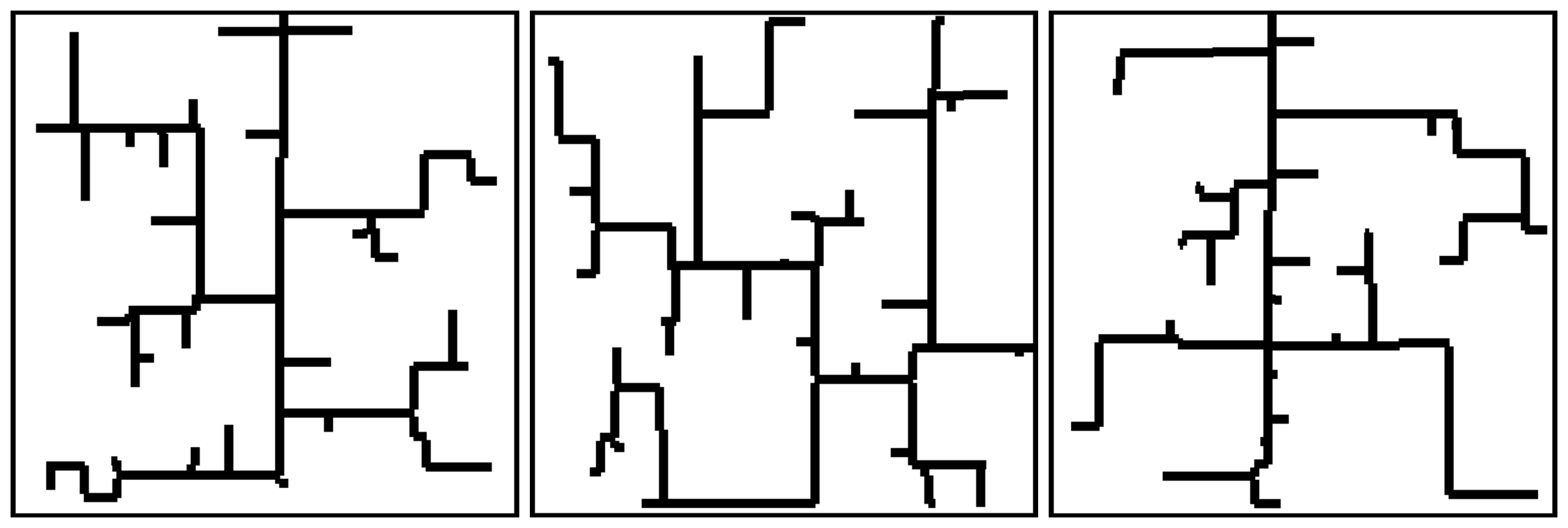}
\caption{Examples of generated random maps}
\label{fig:random-map-examples}
\end{figure}

In the process of generating random maps, a structure resembling the interior layout of office-like buildings is adopted. During this process, each obstacle is placed on the map perpendicular to another obstacle to simulate a multi-room environment. Examples of random maps generated with this strategy are shown in Figure \ref{fig:random-map-examples}.

\begin{figure*}
\centering
\subfloat[Planning time]{%
\label{fig:random-time-graph}
\includegraphics[width=0.32\textwidth]{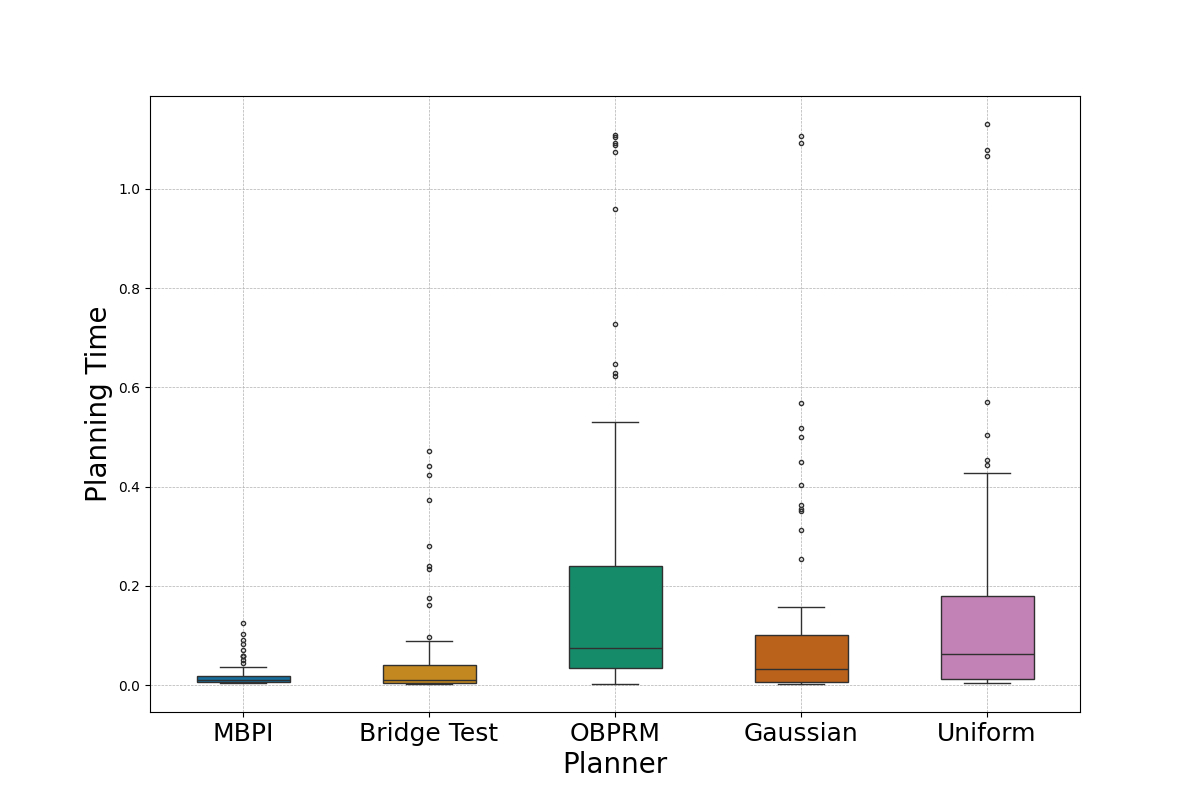}}\hfill
\subfloat[Number of milestones]{%
\label{fig:random-milestone-graph}
\includegraphics[width=0.32\textwidth]{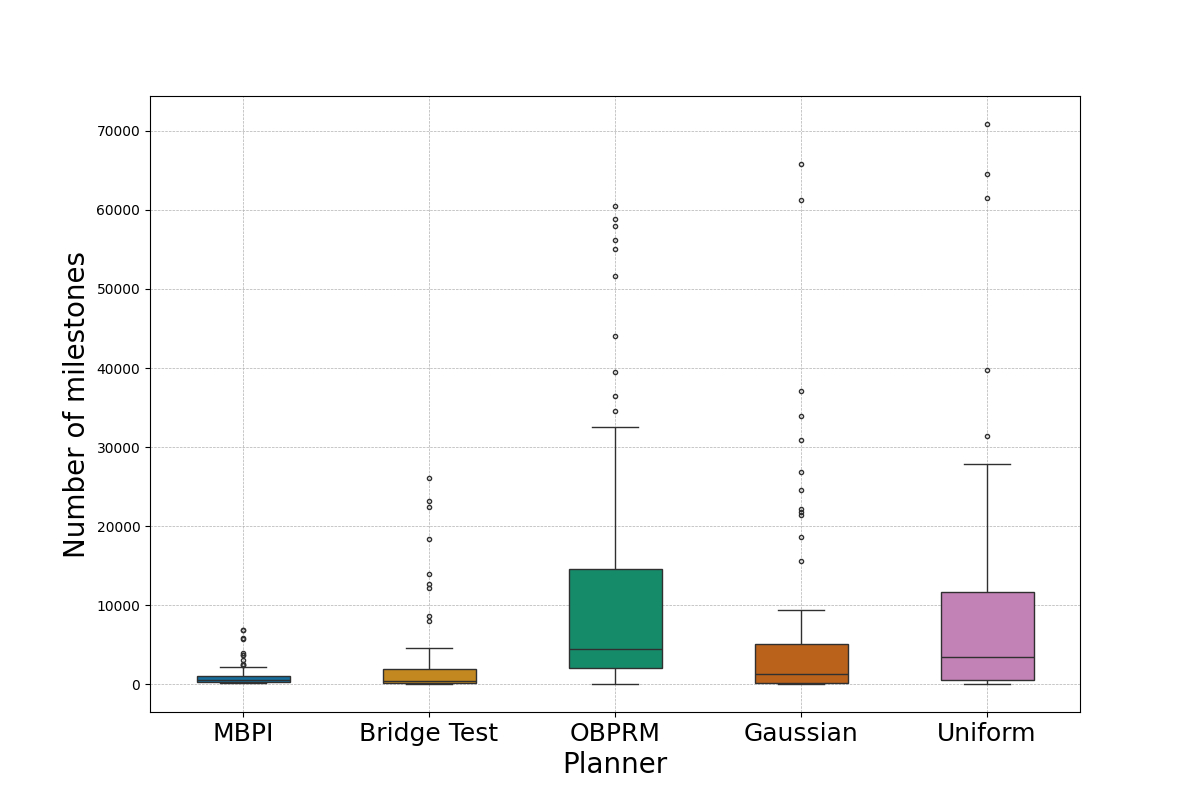}}\hfill
\subfloat[Path length]{%
\label{fig:random-length-graph}
\includegraphics[width=0.32\textwidth]{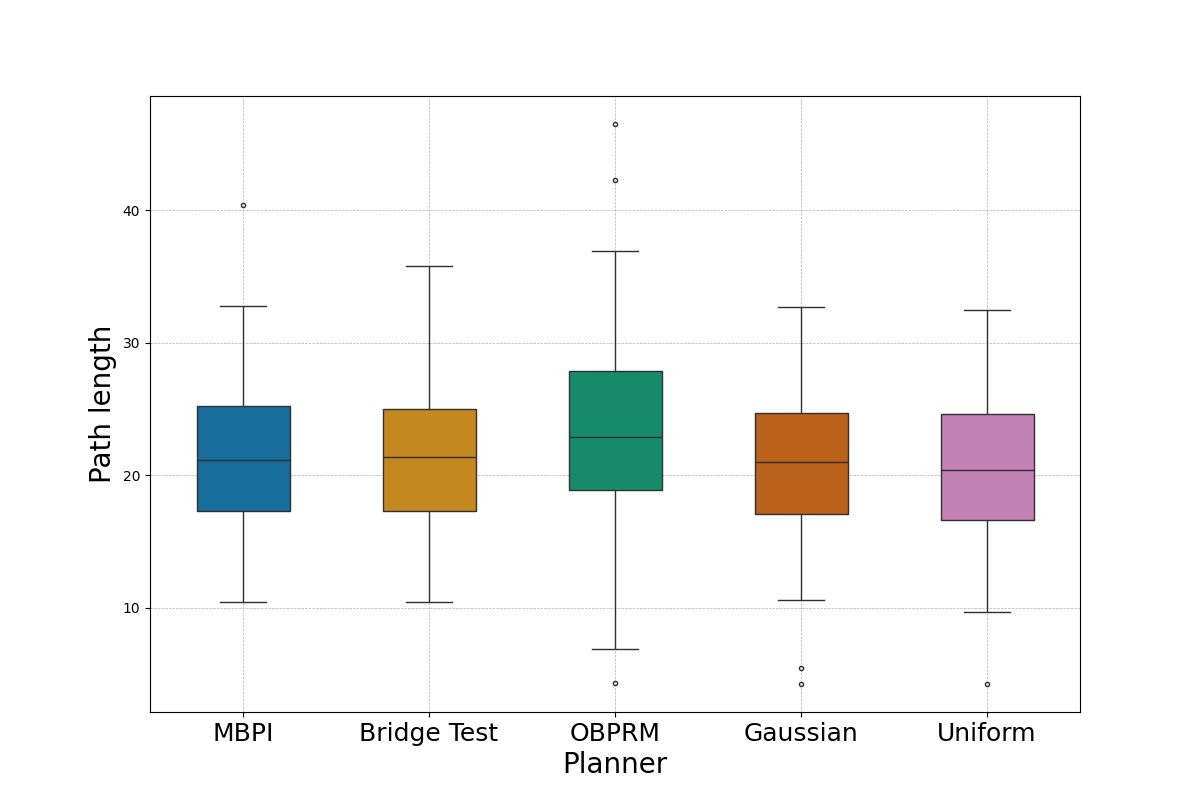}}
\caption{Box plots of random map experiments}
\label{fig:random-results}
\end{figure*}

\begin{table}[ht]
    \centering
    \caption{Random map experiment performances}
    \renewcommand{\arraystretch}{1.5}
    \begin{tabular}{|c|c|c|c|c|}
    \hline
    \multirow{2}{*}{\textbf{Sampler}} & \multicolumn{3}{c|}{\textbf{Metrics} (Avg.(Std. dev.))} \\ 
                            & \multicolumn{1}{c}{\(t_p\)\footnotemark[1]} & \multicolumn{1}{c}{\(n_m\)\footnotemark[2]} & \multicolumn{1}{c|}{\(l_p\)\footnotemark[3]} \\ \hline
    MBPI & $0.018(0.021)$ & $999.1(1353)$ & $21.52(5.60)$ \\ \hline
    Bridge Test & $0.046(0.093)$ & $2265.9(4927)$ & $21.64(5.29)$ \\ \hline
    OBPRM & $0.200(0.274)$ & $11399.0(15003)$ & $23.29(6.85)$ \\ \hline
    Gaussian & $0.099(0.187)$ & $5643.6(11314)$ & $20.77(5.63)$ \\ \hline
    Uniform & $0.143(0.215)$ & $8770.1(13389)$ & $20.75(5.67)$ \\ \hline
    \end{tabular}
    \label{tab:random-results}
    \footnotetext[1]{Planning Time (s)}
    \footnotetext[2]{Number of Milestones}
    \footnotetext[3]{Path Length (m)}
\end{table}

A total of one thousand maps are generated using the defined strategy. The initial and target points of each map are randomly selected, ensuring a minimum distance between them. On these maps, PRM is executed with each sampler. For each path planning process, the planning time, number of milestones, and path length are measured. The results are combined to produce the graphs shown in Figure \ref{fig:random-results}. In these graphs, the results of each sampler for each metric are shown as box plots. In addition, Table \ref{tab:random-results} shows the averages and standard deviations of the results obtained from the experiment.  

According to the results, the performance of MBPI is superior to other methods in terms of average planning time and number of milestones. In addition, the standard deviations of these metrics are lower than the other methods, indicating that MBPI gives more stable results. Furthermore, the performance in path length is near to the other methods.

\subsubsection{Real-world Experiments}\label{sec:real-world-experiments}
To evaluate the practical applicability and effectiveness of the presented work, the algorithm is tested on a robot in a real-world environment. As in previous experiments, existing methods are also tested in this environment to evaluate objectively the performance of MBPI.

\begin{figure}
\centering
\subfloat[Turtlebot3 Burger]{%
\label{fig:turtlebot}
\includegraphics[width=0.25\textwidth]{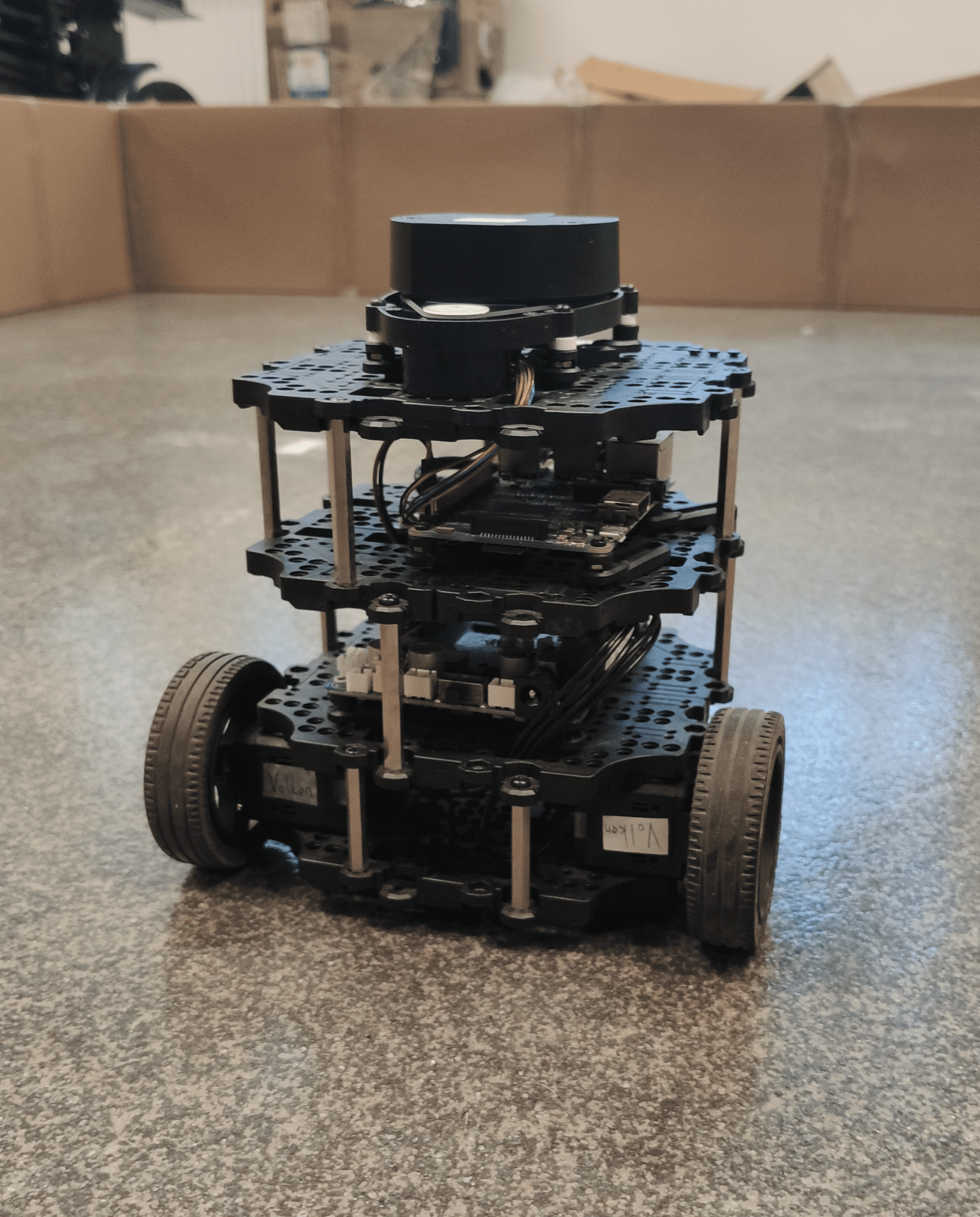}}\hspace{5pt}
\subfloat[Map of the real-world environment generated by the Turtlebot]{%
\label{fig:parkour-pgm}
\includegraphics[width=0.28\textwidth]{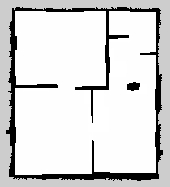}}
\caption{Real-world environment}
\label{fig:real-world-environment}
\end{figure}

The Turtlebot3 Burger \citep{amsters2020turtlebot} shown in Figure \ref{fig:turtlebot} is selected as the robot for real-world experiments. To test the sampler algorithms on the Turtlebot, they must be integrated into the Robot Operating System (ROS) environment. As a result, a new global planner plugin is implemented in the \textit{move\_base} package of ROS. The main role of the plugin is to connect OMPL planners and samplers to ROS. During the experiment, the existing navigation packages of the Turtlebot are utilized by replacing the global planner with this plugin. Furthermore, due to the presence of narrow passages in the environment, it is necessary to have a local planner algorithm that can work efficiently in such areas. After testing various planners, the Teb Local Planner has been found to work effectively in such environments. Therefore, the Teb Local Planner is preferred as the local planner.

\begin{figure}
\centering
\includegraphics[width=12cm]{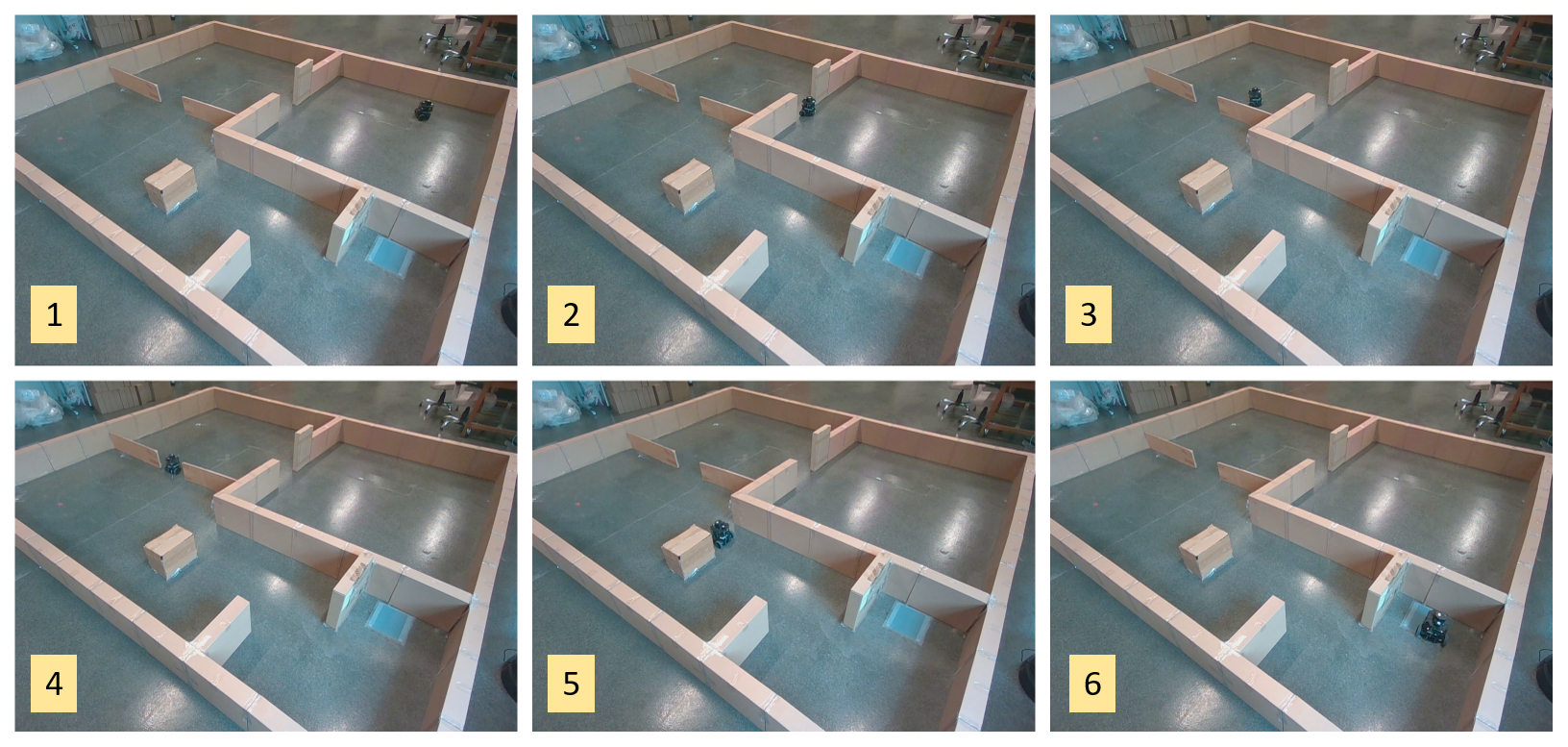}
\caption{Scenes from the real-world experiment}
\label{fig:parkour-collage}
\end{figure}

The experiments are conducted in a static environment that had been previously mapped. The map of the environment scanned by Turtlebot3 Burger is shown in Figure \ref{fig:parkour-pgm}. The experimental process of each sampler is video recorded along with synchronized visualizations from Rviz. The video can be accessed from supplementary files. In Figure \ref{fig:parkour-collage} scenes from the experiment are shown. 

\begin{table}[ht]
    \centering
    \caption{Real-world experiment performances. The combined time for MBPI is given in paranthesis.}
    \renewcommand{\arraystretch}{1.5}
    \begin{tabular}{|c|c|c|c|c|}
    \hline
    \multirow{2}{*}{\textbf{Sampling Method}} & \multicolumn{3}{c|}{\textbf{Metrics} (Avg.)} \\
                               & \multicolumn{1}{c}{\(t_p\)\footnotemark[1]} & \multicolumn{1}{c}{\(n_m\)\footnotemark[2]} & \multicolumn{1}{c|}{\(l_p\)\footnotemark[3]} \\ \hline
    MBPI ($0.0033$) & $0.0022$ & $103.8$ & $6.41$ \\ \hline
    Bridge Test & $0.0043$ & $102.3$ & $6.54$ \\ \hline
    OBPRM & $0.0052$ & $303.9$ & $6.53$ \\ \hline
    Gaussian & $0.0025$ & $94.7$ & $6.45$ \\ \hline
    Uniform & $0.0025$ & $101.7$ & $6.43$ \\ \hline
    \end{tabular}
    \label{tab:real-world-results}
    \footnotetext[1]{Planning Time (s)}
    \footnotetext[2]{Number of Milestones}
    \footnotetext[3]{Path Length (m)}
\end{table}

Using each sampler, planning is performed 5 times between the same start and target points. Table \ref{tab:real-world-results} shows the results for the planning time, number of milestones in the roadmap, and path length results obtained from the experiment. According to the results shown in the table, the algorithms give similar results. MBPI provides the best results in planning time and path length metrics. 

To further analyze the algorithm's performance, the combination of planning and identification times is evaluated in real-world experiments. The combined time is given in Table 4 and calculated using the Equation \ref{eq:combined_time}. In this context, $t_p$ denotes the average planning time, $n_p$ represents the number of planning instances, and $t_i$ corresponds to the identification time.

\begin{align}
\begin{split}
\text{Combined Time} &= \frac{t_pn_p + t_i}{n_p}
\end{split}
\label{eq:combined_time}
\end{align}

This formula highlights that as the number of replanning ($n_p$) attempts increases, the impact of the identification time ($t_i$) becomes negligible. In static environments, the identification algorithm is executed only once, and the resulting probability density function is used repeatedly during replanning.

In dynamic environments, if narrow passages remain unaffected by moving obstacles, a single execution of the identification algorithm suffices, allowing the pre-computed narrow passages to guide the planning process. Additionally, when dynamic obstacles alter the position of narrow passages, the hybrid sampling strategy ensures that feasible paths can still be computed. However, this may reduce planning efficiency due to outdated narrow passages. To overcome this issue, the identification algorithm can be executed at a lower frequency, while replanning operates at a higher frequency to adapt to dynamic environment.

The results of the real-world test are parallel to the results obtained in simulation environments. These results indicate that the algorithm is applicable.

\section{Conclusion and Future Work}\label{sec:conclusion-and-future-work}
This paper introduces an approach to identify narrow passages, showcasing its effectiveness within the PRM framework for generating roadmaps with high connectivity and coverage. Uniform sampler, standard PRM sampler, provides good coverage by sampling entire map with equal probability. However, connectivity problem of roadmaps appears in environments with narrow passages. To solve this problem, MBPI detects and samples narrow passages. Moreover, a uniform sampler is utilized to ensure high coverage of roadmap. To achieve both, narrow passage sampler and uniform sampler are used hybridly. 

MBPI is compared in simulation environment with Bridge Test, Gaussian, Obstacle-based, and Uniform samplers available in OMPL. In addition, algorithms are compared in a real-world environment using Turtlebot3 Burger. The experiments are conducted with metrics of planning time, number of milestones, and path length. The results of the experiments show that MBPI outperforms other methods in terms of planning time and number of milestones, and generally improves path length.

Further studies can be carried out to improve planning strategies in narrow passage environments. These include adapting our method to work with other sampling-based algorithms whose performance is degraded in narrow passage environments. Moreover, broadening the algorithm’s applicability to high-dimensional spaces is expected to expand its utility, enabling the handling of more complex scenarios such as manipulator arm planning. Although main advantage of sampling-based methods is observed in high-dimensional environments, this work is anticipated to inspire future studies on adaptation to high-dimensional spaces and development of sampling-based algorithms. In addition, combining the narrow passage detection algorithm with quadtree \citep{samet1984quadtree} decomposition could offer advantages by providing a more efficient and better map representation. This approach increases likelihood of seeing narrow passages, which are very small.

\section*{Disclosure statement}
The authors have no relevant financial or nonfinancial
interests to disclose.

\section*{Funding}
This work was supported by the Turkish Scientific and Technological Research Council (TUBITAK) under project no. 121E537.

\section*{Notes on contributors}
The study was conceptualized by Y.E.Ş. Methodology design was carried out by Y.E.Ş., E.Y.G., and V.S. Formal analysis and investigation were conducted by Y.E.Ş., E.Y.G., and V.S. The original draft of the manuscript was prepared by Y.E.Ş., E.Y.G., and V.S. V.S. contributed to the review and editing of the manuscript. Additionally, V.S. provided supervision for the project.

\end{document}